\documentclass[preprint,authoryear,12pt]{elsarticle}

\usepackage{amssymb}
\usepackage{amsthm}
\usepackage{amsmath}
\usepackage{url}
\newtheorem{thm}{Theorem}
\newtheorem{cor}{Corollary}
\newtheorem{lem}{Lemma}
\newtheorem{prop}{Proposition}
\theoremstyle{definition}
\newtheorem{defn}{Definition}
\newtheorem{rem}{Remark}
\newtheorem{ex}{Example}
\usepackage{lineno}

\biboptions{longnamesfirst,comma,semicolon}

\journal{Neural Networks}

\begin{document}

\begin{frontmatter}

	\title{Storing Cycles in Hopfield-type Networks with Pseudoinverse Learning Rule: Admissibility and Network Topology}

	\author[csu]{Chuan Zhang}

	\author[csu]{Gerhard Dangelmayr}

	\author[csu]{Iuliana Oprea\corref{cor1}}
	\ead{juliana@math.colostate.edu}
	\cortext[cor1]{Corresponding author}

	\address[csu]{Department of Mathematics, Colorado State University, Fort Collins, CO 80523, USA}

	\begin{abstract}
		Cyclic patterns of neuronal activity are ubiquitous in animal nervous systems, and partially responsible for generating and controlling rhythmic movements such as locomotion, respiration, swallowing and so on. Clarifying the role of the network connectivities for generating cyclic patterns is fundamental for understanding the generation of rhythmic movements. In this paper, the storage of binary cycles in Hopfield-type and other neural networks is investigated. We call a cycle defined by a binary matrix $\Sigma$ admissible if a connectivity matrix satisfying the cycle's transition conditions exists, and if so construct it using the pseudoinverse learning rule. Our main focus is on the structural features of admissible cycles and the topology of the corresponding networks. We show that $\Sigma$ is admissible if and only if its discrete Fourier transform contains exactly $r=\mathrm{rank}(\Sigma)$ nonzero columns. Based on the decomposition of the rows of $\Sigma$ into disjoint subsets corresponding to loops, where a loop is defined by the set of all cyclic permutations of a row, cycles are classified as simple cycles, and separable or inseparable composite cycles. Simple cycles contain rows from one loop only, and the network topology is a feedforward chain with feedback to one neuron if the loop-vectors in $\Sigma$ are cyclic permutations of each other. For special cases this topology simplifies to a ring with only one feedback. Composite cycles contain rows from at least two disjoint loops, and the neurons corresponding to the loop-vectors in $\Sigma$ from the same loop are identified with a cluster. Networks constructed from separable composite cycles decompose into completely isolated clusters. For inseparable composite cycles at least two clusters are connected, and the cluster-connectivity is related to the intersections of the spaces spanned by the loop-vectors of the clusters. Simulations showing successfully retrieved cycles in continuous-time Hopfield-type networks and in networks of spiking neurons exhibiting up-down states are presented.
	\end{abstract}


\end{frontmatter}


\section{Introduction}
Applications of artificial neural networks in content addressable (associative) memory have attracted much attention in the last few decades \citep{Hopfield82, Hopfield84, Little, Capacity, Universal, Associative}. Hopfield-type networks are among the most popular models of artificial neural networks for studying content addressable memory. According to Hopfield's original idea, the privileged regime to store information has been fixed point attractors, however experiments \citep[e.g.,][]{ChaosBrain} indicate that cycles are used to store information and chaotic dynamics appears as the background regime composed of these cyclic ``memory bags''. 

In general, the storage of pattern sequences is one of the most important tasks in both biological and artificial intelligence systems. A sequence containing repetitions of the same subsequence is said to be complex \citep{Guyon,WangArbib90,Wang03}, and cyclic patterns (or cycles of patterns) are one of the important classes of such sequences. In animal nervous systems, cyclic patterns of neuronal activity are ubiquitous and partially responsible for generating and controlling rhythmic movements such as locomotion, respiration, swallowing and so on. Neural networks that can produce cyclic patterned outputs without rhythmic sensory or central input are called central pattern generators (CPGs). While in some lower level invertebrate animals detailed connectivity diagrams among identified CPG neurons have been experimentally determined, the anatomic structure of CPG networks in most higher vertebrate animals including human beings remain largely unknown \citep[e.g.,] [] {MackayLyons02, Marder05, Selverston10}. 

According to Yuste \citep{Yuste08}, the network connectivity problem, i.e. experimentally identifying the connectivity diagram of biological neural networks, is one of the four basic problems that have to be solved to fully understand a biological neural network. However, recent experimental observations \citep[e.g.,][]{Dickinson92,Meyrand94} suggested that CPGs may be highly flexible, some of them may even be temporarily formed only before the production of motor activity \citep{Jean01}. This makes experimentally identifying the architecture of CPGs very difficult. As indirect approaches to solve the network connectivity problem, observable movement features such as symmetry etc. have been used to infer aspects of CPG structures \citep[e.g.][]{Golubitsky99}. In this paper, we study the network connectivity problem for storing binary cyclic patterns. Given an arbitrary binary cyclic pattern, we ask whether there exists a network whose architecture allows to produce it, and if there exists one, then how the cycle determines the network structure. While the main motivation for our study is the storage of cycles in continuous-time Hopfield-type networks, this question is independent of the specific dynamics of the individual neurons the network is composed of.

In both discrete and continuous asymmetric variants of Hopfield-type networks, the storage and retrieval of sequences including cycles of binary patterns have been investigated \citep{Personnaz, Guyon, Gencic90}, and biologically plausible learning rules such as Hebb's rule, the pseudoinverse rule and their variants with and without delays have been used. In this paper, we follow \cite{Gencic90} and use continuous-time Hopfield-type networks as models to study the relation between cyclic patterns and the architecture of the networks constructed from them. In addition to the simple dynamics of single neurons, another advantage of Hopfield-type networks is that they deal with binary states. In neurophysiology it is well known that both CPG neurons and cortical neurons show bistable membrane behaviors, which are commonly referred to as plateau potentials \citep[e.g][]{Straub02,Grillner03,Selverston10} or up-down states \citep[e.g.] [] {Sanchez00, Cossart03}. Accordingly, a sequence of the binary states $+1$ and $-1$ traversed by a single neuron in a Hopfield-type network can be interpreted as a sequence of up and down states, respectively.

While simulations of networks constructed using Hebbian learning rules have been shown to be qualitatively consistent with experimental recordings \citep{Kleinfeld88}, it is well known that Hopfield-type networks with Hebbian learning rules do not perform well when the patterns to be stored are correlated which is usually the case in practice \citep{HebShort1}. To avoid this problem, a pseudoinverse learning rule was introduced by Amari \citep{Amari77}, and in the Hopfield framework by Personnaz et al. and Kanter et al. \citep{Personnaz,Sompolinsky87}. It has been suggested that the pseudoinverse learning rule and its variants may take key roles in the associative perception of human faces in the human cortex \citep{Prosopagnosia00} and the encoding of location information in the rat hippocampus \citep{Marinaro07}. Recently, Tapson and Schaik proposed an algorithm referred to as OPIUM (Online Pseudoinverse Update Method) for computing the pseudoinverse, and showed that the pseudoinverse learning rule is plausible as a physiological process in real neurons \citep{TapsonSchaik13}. Since the pseudoinverse method gives an exact solution of the network connectivity problem if a solution exists \citep{Personnaz}, we use this method to construct networks for storing binary cycles.

Although the pseudoinverse rule and its variants \citep{Amari77,Personnaz} extend to more general cases, most investigations in discrete-time Hopfield-type networks characterized or were implemented for cycles or sequences of linearly independent patterns \citep[e.g.][]{Personnaz,Sompolinsky87}. An approach to storing cycles of correlated as well as linearly independent patterns in continuous-time Hopfield-type networks has been proposed by \cite{Gencic90}. In this study, a successfully retrieved cycle is revealed as an attracting limit cycle in the network dynamics, but the question for which cycles the corresponding network connectivity problem admits a solution was not addressed.

In our study, a cycle is defined by a $N\times p$-matrix, $\Sigma$, of binary states, where $N$ is the number of neurons in the network and $p$ is the length of the cycle. The network connectivity problem associated with a cycle $\Sigma$ can be formulated as follows: Find a real $N\times N$-matrix $\mathbf{J}$ such that $\mathbf{J}\Sigma = F$, where $F$ is related to $\Sigma$ by a cyclic permutation of the columns. The main objective of this paper is to study the existence and properties of solutions of this equation along with the structural features of the corresponding cycles, and the network topologies associated with them. If a solution $\mathbf{J}$ exists, we call the cycle $\Sigma$ admissible and construct $\mathbf{J}$ using the pseudoinverse method.

While the main motivation for our study is the storage of binary cycles in continuous-time Hopfield-type neural networks, the question whether a given cycle is admissible is independent of the particular network-model. For the discrete-time Hopfield-type networks studied by \cite{Personnaz} and \cite{Guyon}, $\mathbf{J}$ can be used directly as connectivity matrix. For the continuous-time Hopfield-type networks considered in Section 2.1, we follow the approach of \cite{Gencic90} and represent the connectivity matrix as a weighted sum of $\mathbf{J}$ and another matrix $\mathbf{J^0}$, that serves to store the individual patterns in $\Sigma$ as fixed points. To demonstrate that our approach also works for more complicated neuron-models, we introduce in Section 2.2 a single-compartment neuron model that exhibits up-down states and show an example of a successfully retrieved cycle.

Our main objective is to analyze and classify the structural features of admissable cycles $\Sigma$ and the topology of the networks constructed from them. A basic result is that if and only if the discrete Fourier transform of $\Sigma$ contains exactly $r=\mathrm{rank}(\Sigma)$ nonzero columns, then $\Sigma$ is admissible.

Our approach to classify cycles is based on the decomposition of the row vectors of $\Sigma$ into disjoint subsets corresponding to different loops created by cyclic permutations of the rows. If the cycle is admissible, each of these loops is associated with an invariant subspace of the row space, $\mathcal{U}$, of $\Sigma$ under cyclic permutations. This row-decomposition leads naturally to a classification of cycles into simple cycles, separable composite cycles, and inseparable composite cycles. Simple cycles contain rows from a single loop only. Composite cycles contain rows from at least two disjoint loops, and for each loop the neurons corresponding to the loop vectors in $\Sigma$ are identified with a cluster, which in turn corresponds to an indecomposable invariant subspace of $\mathcal{U}$ under cyclic permutations if $\Sigma$ is admissible. Two clusters are directly connected if their subspaces intersect nontrivially, and they are connected if they are part of a chain of directly connected clusters. A network constructed from a simple admissible cycle has only one cluster, and we show that the network topology is a feedforward chain with feedback to one neuron if the loop vectors in $\Sigma$ are all cyclic permutations of each other. For special simple cycles, there is only one feedback and the network topology simplifies to a ring. Networks constructed from separable composite cycles decompose into completely isolated clusters.

The paper is organized as follows. In Section 2, the pseudoinverse learning rule is introduced in the framework of continuous-time Hopfield-type networks. Additionally, in order to demonstrate that the pseudoinverse learning rule can be applied to other networks as well, networks of spiking neurons with plateau membrane potentials and postinhibitory rebound are introduced, and simulations showing the successful retrieval of a prescribed cycle are presented. In Section 3, the general admissibility criterion in terms of the discrete Fourier transform of the cycle matrix is formulated and proved, and the relation of admissible cycles with cyclic permutation groups is discussed. Based on the structural features of the invariant subspaces of the row space of an admissible cycle, in Section 4, admissible cycles are classified into simple cycles, and separable and inseparable composite cycles, and for each type of cycles a corresponding admissibility condition is derived. In Section 5, the topologies of networks constructed from different types of admissible cycles are studied, and in Section 6 implications of the results presented in this paper are discussed.

\section{Pseudoinverse Learning Rule and Neural Networks}
\subsection{Hopfield-type Neural Networks}
A continuous-time Hopfield-type network \citep{Hopfield84} is described by a system of ordinary differential equations for $u_i(t)$, $1\leq i\leq N$, which model the membrane potential of the $i$-th neuron in the network at time $t$. Assuming that all neurons are identical, normalizing the neuron amplifier input capacitance and resistance to unity and neglecting external inputs, the governing equations are,
\begin{equation}
	\label{eq:HopfieldNetworkDefnScalar}
	\frac{du_i}{dt} = -u_i + \sum\limits_{j=1}^{N}\tilde{J}_{ij}v_j\mbox{, }1\leq i\leq N,
\end{equation}
where $v_j(t)$ is the firing rate of the $j$-th neuron and $\mathbf{\tilde{J}} = (\tilde{J}_{ij})_{N\times N}$ is the connectivity matrix. The firing rate $v_j$ is related to the membrane potential $u_j$ through a sigmoid-shaped gain function, $v_j = g(u_j)$, which we choose, following \cite{Hopfield84}, as $g(u_j) = \tanh(\lambda u_j)$, where $\lambda$ controls the steepness. Using vector notation, $\mathbf{u} = (u_1,\dots,u_N)^T$, $\mathbf{v} = (v_1,\dots,v_N)^T$, (\ref{eq:HopfieldNetworkDefnScalar}) can be more compactly written as (dots denote time derivatives)
\begin{equation}
	\label{eq:HopfieldNetworkDefnVector}
	\dot{\mathbf{u}} = -\mathbf{u} + \tilde{\mathbf{J}}\tanh(\lambda\mathbf{u}),
\end{equation}
where here and subsequently a scalar function applied to a vector or matrix denotes the vector or matrix obtained by applying the function to each component, i.e.
$$
	\tanh(\lambda\mathbf{u}) = (\tanh(\lambda u_1),\tanh(\lambda u_2),\dots,\tanh(\lambda u_N))^T.
$$
Alternatively, since $u_j = \mathrm{arctanh}(v_j)/\lambda$, (\ref{eq:HopfieldNetworkDefnVector}) can be rewritten as a system of differential equations for the firing rates,
\begin{equation}
	\label{eq:HopfieldNetworkDefnVectorFiringRate}
	\dot{\mathbf{v}} = \lambda(I - \mathrm{diag}(\mathbf{v}^2))(\tilde{\mathbf{J}}\mathbf{v} - \frac{\mathrm{arctanh}(\mathbf{v})}{\lambda}),
\end{equation}
where $I$ is the $N\times N$ identity matrix.

In this paper we study the structure of binary pattern cycles that can be stored in the network modeled by the above autonomous system. Following \cite{Hopfield82,Hopfield84}, any $N$-dimensional $\{-1,1\}$-valued column vector is identified with a \emph{binary vector} or \emph{pattern}, and we use $+$ and $-$ to denote $1$ and $-1$.

	\cite{Personnaz} and \cite{Guyon} studied the storage of sequences of patterns in discrete-time Hopfield-type networks. A sequence of $p$ patterns $\xi^{(\mu)} = (\xi_1^{(\mu)},\xi_2^{(\mu)},\dots,\xi_N^{(\mu)})^T$, $1\leq\mu\leq p$, $\xi_i^{(\mu)} = +\mbox{ or }-$, is defined by $p$ transition conditions $\xi^{(\mu)}\rightarrow f^{(\mu)}$, where $f^{(\mu)}$ is one of the given vectors, i.e., $f^{(\mu)}\in\{\xi^{(1)},\dots,\xi^{(p)}\}$ for each $\mu$. Thus a sequence is characterized by two $N\times p$-matrices $\Sigma = (\xi^{(1)},\dots,\xi^{(p)})$ and $F = (f^{(1)},\dots,f^{(p)})$. The two matrices are related to each other by the transition conditions, which can be conveniently formulated in terms of a $p\times p$ transition matrix as $F = \Sigma\mathbf{P}$, where $\mathbf{P}_{\nu\mu} = 1$ if $f^{(\mu)} = \xi^{(\nu)}$ and $0$ otherwise. For example, for $p=3$ and the simplest case of a sequence starting at $\xi^{(1)}$ and terminating at $\xi^{(3)}$, $\xi^{(1)}\rightarrow\xi^{(2)}$, $\xi^{(2)}\rightarrow\xi^{(3)}$, $\xi^{(3)}\rightarrow\xi^{(3)}$, we have $F = (\xi^{(2)},\xi^{(3)},\xi^{(3)})$ and $P$ is singular, but the general definition in terms of $\Sigma$ and $F$ allows to consider more complex as well as multiple sequences. \cite{Personnaz} showed that the storage of such a sequence leads to the matrix equation,
\begin{equation}
	\label{eq:Transition00}
	\mathbf{J}\Sigma = F,
\end{equation}
for the connectivity matrix $\mathbf{J}$ of the discrete network. It was pointed out by \cite{Personnaz}, that, if $F\Sigma^+\Sigma = F$, where $\Sigma^+$ is the Moore-Penrose pseudoinverse of $\Sigma$, then (\ref{eq:Transition00}) has the exact solution $\mathbf{J} = F\Sigma^+$, which was called \emph{associating learning rule} by these authors.

A \emph{cycle} of $p$ patterns is a sequence with $F = (\xi^{(2)},\xi^{(3)},\dots,\xi^{(p)},\xi^{(1)})$, i.e., $\xi^{(\mu)}\rightarrow\xi^{(\mu+1)}$ for $\mu<p$ and $\xi^{(p)}\rightarrow\xi^{(1)}$, and the corresponding transition matrix is
\begin{equation}
	\label{eq:Def_P00}
	\mathbf{P} = \left(\begin{array}{rrrrrr}
		 0 &  0 &  0 & \cdots &  0 &  1 \\
		 1 &  0 &  0 & \cdots &  0 &  0 \\
		 0 &  1 &  0 & \cdots &  0 &  0 \\
		\vdots & \vdots & \vdots & \ddots & \vdots & \vdots \\
		 0 &  0 &  0 & \cdots &  0 &  0 \\
		 0 &  0 &  0 & \cdots &  1 &  0 
	\end{array}\right).
\end{equation}
We are interested in the storage of cycles in the continuous-time Hopfield networks defined by (\ref{eq:HopfieldNetworkDefnVector}). Our approach to compute a connectivity matrix $\mathbf{\tilde{J}}$ for this purpose follows \cite{Gencic90}. In this paper, the connectivity matrix $\mathbf{\tilde{J}}$ is decomposed as
\begin{equation}
	\label{eq:Connection00}
	\mathbf{\tilde{J}} = \beta_K(C_0\mathbf{J^0} + C_1\mathbf{J}),
\end{equation}
where $\mathbf{J^0}$ serves to stabilize the network in its current memory state and $\mathbf{J}$ imposes the transitions between the memory states. Here, $C_1 = 1 - C_0$ and $C_0$, $0\leq C_0\leq 1$, control the relative contributions of the two components of $\mathbf{\tilde{J}}$. The fixed point condition is realized by requiring that $\mathbf{v} = \beta_1\xi^{(\mu)}$, with a parameter $0< \beta_1 <1$, is a fixed point if $C_0 = 1$. Noting that $\mathrm{arctanh}(x)$ is an odd function and $|\xi^{(\mu)}_i| = 1$, this leads, according to (\ref{eq:HopfieldNetworkDefnVectorFiringRate}), to the condition
$$
	\mathbf{\tilde{J}}\beta_1\xi^{(\mu)} = \frac{1}{\lambda}\mathrm{arctanh}(\beta_1\xi^{(\mu)}) = \frac{1}{\lambda}\mathrm{arctanh}(\beta_1)\xi^{(\mu)},
$$
for every $\mu$, hence $\mathbf{\tilde{J}}\Sigma = \beta_K\Sigma$ with $\displaystyle{\beta_K = \frac{1}{\lambda\beta_1}\mathrm{arctanh}(\beta_1)}$, which has the solution $\mathbf{\tilde{J}} = \beta_K\mathbf{J^0}$ with
\begin{equation}
	\label{eq:FPCond00}
	\mathbf{J^0} = \Sigma\Sigma^+.
\end{equation}
Regarding the transition conditions, we stipulate that $\mathbf{v}(t) = \beta_1\xi^{(\mu)}$ implies $\mathbf{v}(t+\tau) = \beta_1 f^{(\mu)}$ for some $\tau$, and require accordingly for $C_0 = 0$ that $\mathbf{\tilde{J}}\Sigma = \beta_K F$. This leads to equation (\ref{eq:Transition00}), which in terms of the transition matrix $\mathbf{P}$, equation (\ref{eq:Def_P00}), becomes
\begin{equation}
	\label{eq:Transition01}
	\mathbf{J}\Sigma = \Sigma\mathbf{P}.
\end{equation}
According to the associating learning rule of \cite{Personnaz}, (\ref{eq:Transition01}) has the solution
\begin{equation}
	\label{eq:Transition02}
	\mathbf{J} = \Sigma\mathbf{P}\Sigma^+,
\end{equation}
provided that $\Sigma\mathbf{P}\Sigma^+\Sigma = \Sigma\mathbf{P}$. If this condition is not satisfied, (\ref{eq:Transition01}) has no solution.

\begin{figure}
	\label{Fig1}
	\begin{center}
		\includegraphics[height=2in]{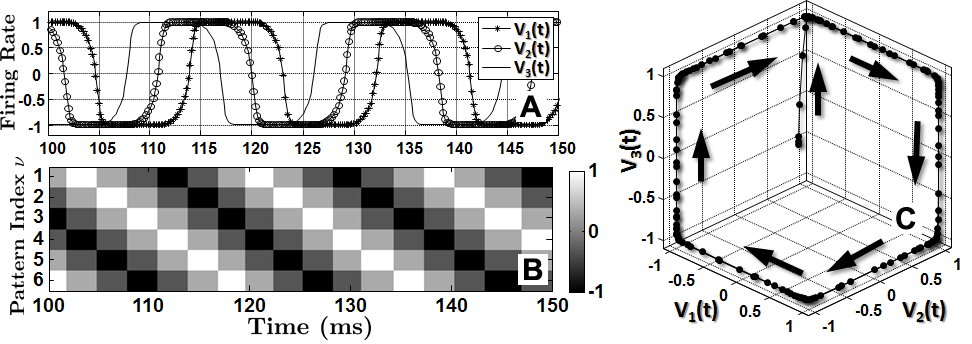}
	\end{center}
	\caption{A successfully retrieved cycle $\Sigma$ (see text for details) in a network of three neurons. \textbf{A} shows the firing rates $v_i(t)$ of the three neurons and the raster plot \textbf{B} shows the overlaps $m^{(\nu)}(t)$, which measure the similarity of the network state $\mathbf{v}(t)$ with each of the six patterns in the cycle (see text). \textbf{C} illustrates the retrieval of the cycle in the phase space of the system.}
\end{figure}

The main objective of this paper is the study of the existence and properties of the solutions of (\ref{eq:Transition01}) along with the structural features of the corresponding cycles, and the network topologies resulting from $\mathbf{J^0}$ and $\mathbf{J}$ defined by (\ref{eq:FPCond00}) and (\ref{eq:Transition02}). A bifurcation analysis and a study of the dynamics of (\ref{eq:HopfieldNetworkDefnVector}) with (\ref{eq:Connection00}), with $C_0$ and $\beta = \lambda\beta_K > 1$ treated as parameters, are given elsewhere, where we also consider the extension of (\ref{eq:HopfieldNetworkDefnVector}) to a dynamical system with a delay,
\begin{equation}
	\label{eq:HopfieldNetwork05}
	\dot{\mathbf{u}} = -\mathbf{u} + C_0\beta_K\mathbf{J^0}\tanh(\lambda\mathbf{u}) + C_1\beta_K\mathbf{J}\tanh(\lambda\mathbf{u}_{\tau}),
\end{equation}
with a delay-time $\tau>0$ and $\mathbf{u}_{\tau}(t) = \mathbf{u}(t - \tau)$. Here we show only one example of a successfully retrieved cycle for (\ref{eq:HopfieldNetworkDefnVector}), a network of $N=3$ neurons. The cycle consists of six states, $\Sigma = (\xi^{(1)},\dots,\xi^{(6)})$, with $\xi^{(1)} = (+,+,+)^T$, $\xi^{(2)} = (+,+,-)^T$, $\xi^{(3)} = (+,-,-)^T$ and $\xi^{(3+\mu)} = -\xi^{(\mu)}$ for $\mu = 1, 2, 3$. The retrieval of the cycle is illustrated in Figure 1. The raster plot \textbf{B} in this figure shows the overlaps, defined in general as
\begin{equation}
	\label{eq:OverlapHopfield}
	m^{(\nu)}(t) = \frac{1}{N}\sum\limits_{i=1}^N v_i(t)\xi_i^{(\nu)}\mbox{, }1\leq\nu\leq p,
\end{equation}
of the actual network state $\mathbf{v}(t)$ with the patterns of the cycle. The overlap $m^{(\nu)}(t)$ is a normalized measure of the similarity of $\mathbf{v}(t)$ with $\xi^{(\nu)}$. Maximal similarity with $\xi^{(\nu)}$ and $-\xi^{(\nu)}$ occurs for $m^{(\nu)}$ close to $1$ and $-1$, respectively. The raster plot of the overlaps in Figure 1B as well as the time series in Figure 1A clearly illustrate that the cycle is retrieved successfully. The parameters $C_0$ and $\beta$ used in this simulation were $C_0 = 0.6$ and $\beta = 4$.

\subsection{Networks of Spiking Neurons}
As was pointed out in the introduction, although our results are developed in the framework of Hopfield-type networks, they also can be used to store cycles in other neural networks. In this subsection, we introduce a network model of identical spiking neurons with bistable membrane behavior and postinhibitory rebound, and show an example of a successfully retrieved cycle in a network constructed using the pseudoinverse method.

We consider the simplest single-compartment neuron model, the passive integrate-and-fire (PIF) model \citep{DayanAbbott01, Abbott07}. The model is described by the following first-order nonlinear ordinary differential equation,
\begin{equation}
	\label{eq:PIFNeuron}
	c_m\frac{dV_i}{dt} = -I_L^{(i)}(t) - I_{nl}^{(i)}(t) + I_{_{synE}}^{(i)}(t) + I_{_{synI}}^{(i)}(t),
\end{equation}
where $V_i(t)$ is the membrane potential of the $i$-th neuron in the network, and if $V_i(t) \geq \theta$ where $\theta = -45\mathrm{mV}$ is the threshold for the firing action potentials, then $V_i(t + dt) = 0$, and $V_i(t + 2dt) = V_{reset}$ with $dt = 0.005\mathrm{ms}$ and $V_{reset} = -55\mathrm{mV}$. After each action potential, an absolute refractory period $t_{Refr} = 1\mathrm{ms}$ is imposed, and during the refractory period the membrane potential $V_i(t)$ is fixed at $V_{reset}$. The parameter $c_m$, chosen as $c_m = 20\mathrm{nF/mm^2}$, is the specific membrane capacitance. The membrane and synaptic currents of the $i$-th neuron are respectively given as follows.

\textbf{\textit{Leakage membrane current:}}
\begin{equation}
	\label{eq:LeakageCurrent}
	I_L^{(i)}(t) = g_L(V_i(t) - E_L),
\end{equation}
where $g_L = 1\mathrm{\mu S/mm^2}$ and $E_L = -68\mathrm{mV}$.

\textbf{\textit{Nonlinear membrane current:}}
\begin{equation}
	\label{eq:NonlinearCurrent}
	I_{nl}^{(i)}(t) = g_{nl}(V_i(t) - E_1)(V_i(t) - E_2)(V_i(t) - E_3) - \delta,
\end{equation}
where $g_{nl} = 0.03\mathrm{mV^{-2}}$, $E_1 = -72\mathrm{mV}$, $E_2 = -58\mathrm{mV}$, $E_3 = -44\mathrm{mV}$ and $\delta$ is a parameter for shifting the nonlinear membrane current to control the stability of the up state. In the simulation shown in the paper, we chose $\delta = -31.685$.

\textbf{\textit{Excitatory synaptic current:}}
\begin{equation}
	\label{eq:ExcitatorySynapticCurrent}
	I_{_{synE}}^{(i)}(t) = \bar{g}_{_{synE}} s_i(t) (V_i(t) - E_{_{synE}}),
\end{equation}
where $\bar{g}_{_{synE}} = 68\mathrm{\mu S/mm^2}$, $E_{_{synE}} = -120\mathrm{mV}$, and the activation variable $s_i(t)$ satisfies the following first order differential equation,
\begin{equation}
	\label{eq:ExcitActivatVariable}
	\frac{ds_i}{dt} = \alpha_E\Big{(}s_i - 10\sum\limits_{j=1}^N \Theta(J_{ij}) J_{ij} \Theta(V_j(t - \tau) - V_{thr})\Big{)},
\end{equation}
where $\alpha_E = 1$, $V_{thr} = -45\mathrm{mV}$, $\tau = 10\mathrm{ms}$, $\Theta(x)$ is the Heaviside step function, and the $J_{ij}$ are the components of the connectivity matrix $\mathbf{J}$.

\textbf{\textit{Inhibitory synaptic current:}}
\begin{equation}
	\label{eq:InhibitorySynapticCurrent}
	I_{_{synI}}^{(i)}(t) = \bar{g}_{_{synI}} z_i(t) (V_i(t) - E_{_{synI}}),
\end{equation}
where $\bar{g}_{_{synI}} = 118\mathrm{\mu S/mm^2}$, $E_{_{synI}} = -120\mathrm{mV}$, and the activation variable is given by $z_i(t) = (x_i(t) + y_i(t))/2$, with $x_i(t)$ and $y_i(t)$ satisfying the following first order differential equations,
\begin{equation}
	\label{eq:InhibitActivatVariableXY}
	\begin{array}{l}
		\displaystyle{\frac{dx_i}{dt}} = \alpha_I\Big{(}x_i - 10\sum\limits_{j=1}^N \Theta(-J_{ij}) J_{ij} \Theta(V_j(t - \tau) - V_{thr})\Big{)}, \\
		\displaystyle{\frac{dy_i}{dt}} = \beta_I\Big{(}y_i + 10\sum\limits_{j=1}^N \Theta(-J_{ij}) J_{ij} \Theta(V_j(t - \tau) - V_{thr})\Big{)},
	\end{array}
\end{equation}
with $\alpha_I = 2$, $\beta_I = 0.08$, and $\tau = 10\mathrm{ms}$.

With the parameters of a single neuron fixed as above, the dynamics of a network of $N$ PIF-neurons is fully determined by the connectivity matrix $\mathbf{J}$. We constructed $\mathbf{J}$ from prescribed cycles using the pseudoinverse learning rule $\mathbf{J} = \Sigma\mathbf{P}\Sigma^{+}$, i.e. without invoking a fixed point condition. Figure 2 illustrates a successfully retrieved $6\times 8$ cycle $\Sigma$. The first four rows of $\Sigma$ are $\sigma_1\mathbf{P}^{j-1}$, $j = 1, 2, 3, 4$, and the last two rows are $\sigma_2$ and $\sigma_2\mathbf{P}$, where $\sigma_1 = (+,+,+,+,-,-,-,-)$ and $\sigma_2 = (+,+,-,-,+,+,-,-)$.

Figure 2A shows the retrieved traces of the membrane potentials $V_i(t)$ of the six neurons in the network. Since the firing rates are not included as variables in the model, they have to be extracted from the time series. Following \cite{DayanAbbott01}, we counted for given $t$ the number of times $t'$ within the time window $t-\Delta t/2 \leq t' \leq t+\Delta t/2$ at which neuron $i$ fired, and divided this number by $\Delta t$. The resulting function, $R_i(t)$, is considered as an approximation of the firing rate of the $i$-th neuron. For $\Delta t$ we chose $\Delta t = 5\mathrm{ms}$. We also introduce the normalized firing rates, $v_i(t) = 2R_i(t)/\max R_i(t) - 1$ (so that $-1\leq v_i(t)\leq 1$, analogous to the firing rates used in continuous-time Hopfield-type networks), and define the overlaps $m^{(\nu)}(t)$ as in equation (\ref{eq:OverlapHopfield}).

To compare the membrane potentials with the prescribed cycle, we extracted the time spans between the first spike and the last spike in each up-state, and identified their average divided by 4 as the time span for each binary state. The resulting time span is $11.3\mathrm{ms}$ and is slightly larger than the time-delay $\tau= 10\mathrm{ms}$ in the synaptic couplings. In Figure 2A, the gray strips in the background indicate these time spans, and the dark gray $+$'s and $-$'s label the corresponding binary states in the prescribed cycle. The firing rates $R_i(t)$ and the overlaps $m^{(\nu )}(t)$ are displayed in Figure 2B and C, respectively. The black arrows in A, B, and C indicate the time span when the first binary pattern, $\xi^{(1)} = (+,+,+,+,+,+)^T$, in the prescribed cycle is retrieved for the first time in the displayed time range. The plots in Figure 2 clearly demonstrate that the cycle is retrieved successfully.

\begin{figure}
	\label{Fig2}
	\begin{center}
		\includegraphics[width=5.8in]{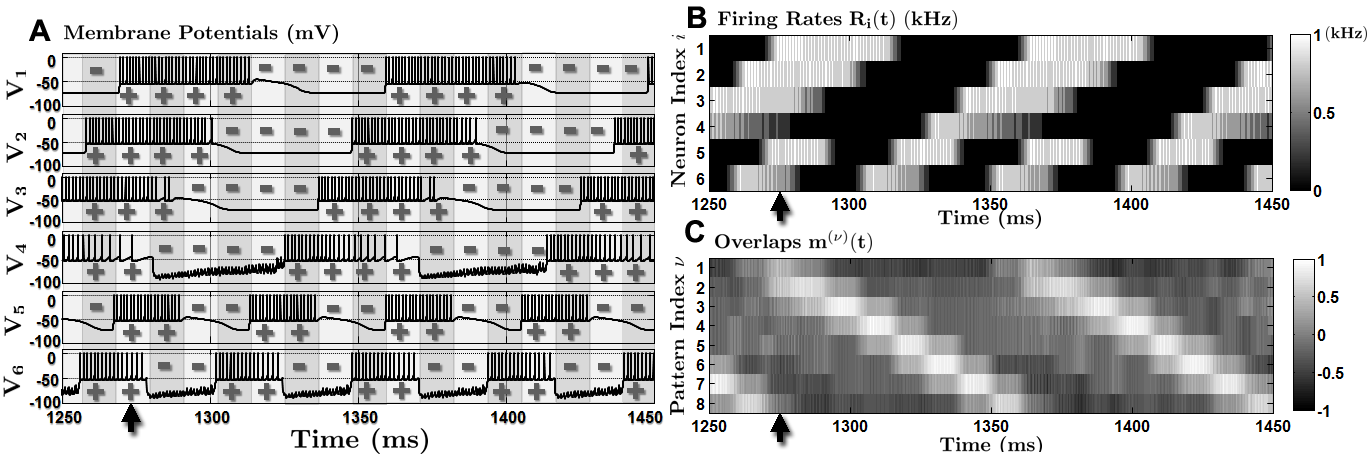}
	\end{center}
	\caption{Successful retrieval of a cycle $\Sigma$ with eight states prescribed in a network of six spiking neurons with the pseudoinverse learning rule. \textbf{A}: membrane potential, \textbf{B}: Firing rates, and \textbf{C}: Overlaps (see text for details).}
\end{figure}

Other cycles were retrieved successfully as well, but in contrast to continuous time Hopfield-type networks, especially with delayed couplings, we observed in simulations that some prescribed cycles are difficult to be retrieved in networks of the spiking neurons introduced in this subsection. This is likely because of the complicated dynamics of the individual neurons, which makes the appropriate choice of parameter values more difficult. In general, especially in physiologically based neural network models, the neuronal dynamics may take key roles in shaping the dynamics of the networks, and reinforce or weaken the contribution of the network structure in reproducing prescribed cycles. In this case, it is important to find out whether a cyclic patterned output in a system arises from a network-based mechanism or not, and if it does, then to which extent and how the cyclic patterned output is determined by the network architecture. 

In the next three sections, without considering the specific dynamics of single neurons, we formulate and prove conditions for any cyclic pattern under which a connectivity matrix in accordance with the cycle's transition conditions can be constructed, and for cycles for which this is the case we analyze and classify their structural features and their relation to the network topology.

\section{Admissible Cycles and Cyclic Permutation Groups}
\defn Let $\mathbf{P}$ be the cyclic $p\times p$-permutation matrix defined in (\ref{eq:Def_P00}). A cycle defined by a binary $N\times p$-matrix $\Sigma = (\xi^{(1)},\xi^{(2)},\dots,\xi^{(p)})$ is said to be \emph{admissible}, if there is a real $N\times N$ matrix $\mathbf{J}$ such that equation (\ref{eq:Transition01}) is satisfied.

Note that if $\Sigma$ is admissible, the solution to (\ref{eq:Transition01}) may be not unique. If there are several solutions, we select (\ref{eq:Transition02}) as distinguished solution because of its close relationship to $\mathbf{J^0}$, see Remark 1(b) below.

\cite{Gencic90} consider a special type of cycles defined by $p$ binary vectors $\Sigma' = (\xi^{(1)},\xi^{(2)},\dots,\xi^{(p)})$, which satisfy the transition condition $\xi^{(1)}\rightarrow\xi^{(2)}\rightarrow\cdots\xi^{(p)}\rightarrow -\xi^{(1)}\rightarrow -\xi^{(2)}\rightarrow\cdots -\xi^{(p)}\rightarrow\xi^{(1)}$. In this paper we consider these cycles as special cases of cycles of period $2p$ with $\Sigma = (\Sigma',-\Sigma')$.

For storing sequences, \cite{Personnaz} pointed out that, if the associating learning rule $F\Sigma^+\Sigma = F$ is satisfied, the rows of $F$ are linear combinations of the rows of $\Sigma$. This follows from the fact that $\Sigma^+\Sigma$ is the orthogonal projection matrix onto the subspace of $\mathbb{R}^p$ spanned by the rows of $\Sigma$. For storing single cycles, their conclusion can be reformulated geometrically as follows:

\prop A cycle $\Sigma$ of size $N\times p$ is admissible, if and only if its row space is invariant under $\mathbf{P}$, i.e.
\begin{equation}
	\label{eq:GenAdCondition00}
	\mathrm{span}\{\mathrm{R}(\Sigma)\} = \mathrm{span}\{\mathrm{R}(\Sigma\mathbf{P})\},
\end{equation}
where $\mathrm{R}(\Sigma)$ denotes the set of all row vectors of $\Sigma$. 

\rm Note that $|\mathrm{R}(\Sigma)| \leq N$, and if $|\mathrm{R}(\Sigma)| < N$ then two or more different rows of $\Sigma$ are identical, that is, the corresponding neurons traverse the same cycle. Although this is a kind of redundancy, we do not exclude this possibility in our general discussion.

We next formulate a useful alternative admissibility criterion involving the eigenspaces of $\mathbf{P}$. Since $\mathbf{P}$ is a circulant matrix, $\mathbf{P}$ has the orthogonal eigenvectors $v^{(k)} = (1,\rho^k,\rho^{2k},\dots,\rho^{(p-1)k})^T$ for $0 \leq k < p$, where $\rho = e^{2\pi\mathbf{i}/p}$ ($\mathbf{i} = \sqrt{-1}$) is the basic primitive $p$-th root of unity \citep{MatrixTheor}, and in this special case the eigenvalues are $\bar{\rho}^k$. We set $V = (v^{(0)}, v^{(1)}, \dots, v^{(p-1)})$, $\lambda_j = \bar{\rho}^{j-1}$ $(1\leq j\leq p)$, $\Lambda = \mathrm{diag}(\lambda_1,\dots,\lambda_p)$ and note that $\mathbf{P} = V\Lambda V^{-1}$ with $V^{-1} = V^*/p$, where here and subsequently complex conjugation is marked by an overbar and an asterisk denotes the adjoint (complex conjugate transpose) matrix or vector.

Since the transition matrix $\mathbf{P}$ leaves its eigenspaces invariant, it follows that if the row space of $\Sigma$ coincides with the direct sum of its projections onto the eigenspaces of $\mathbf{P}$, then $\Sigma$ is admissible. Based on this consideration, we obtain the following admissibility criterion.

\thm Let $\Sigma$ be a cycle whose matrix form is of size $N\times p$, and let $\hat{\Sigma} = \Sigma V$. Then $\Sigma$ is admissible, if and only if $\hat{\Sigma}$ has precisely $r$ nonzero columns, where $r = \mathrm{rank}(\Sigma) = \mathrm{rank}(\hat{\Sigma})$.
\begin{proof}
	Noting that $\mathbf{J}\Sigma=\Sigma\mathbf{P}=\Sigma V\Lambda V^{-1}$ implies $\mathbf{J}\hat{\Sigma} = \hat{\Sigma}\Lambda$, it follows that $\Sigma$ is admissible if and only if there exists an $N\times N$-matrix $\mathbf{J}$ such that
	\begin{equation}
		\label{eq:AdmissibilityCond00}
		\mathbf{J}\hat{\Sigma} = \hat{\Sigma}\Lambda,
	\end{equation}
	which implies
	\begin{equation}
		\label{eq:AdmissibilityCond01}
		\mathbf{J}\mathrm{col}_j(\hat{\Sigma}) = \lambda_j\mathrm{col}_j(\hat{\Sigma})\mbox{, }1\leq j\leq p,
	\end{equation}
	where $\mathrm{col}_j(\hat{\Sigma})$ denotes the $j$-th column of $\hat{\Sigma}$.

	Suppose now that $\Sigma$ is admissible and $\mathrm{rank}(\Sigma) = r$. Since the columns of $V$ consist of eigenvectors associated to distinct eigenvalues of $\mathbf{P}$, it follows that $\hat{\Sigma} = \Sigma V$ has $r$ linearly independent columns. Assume $\mathrm{col}_{\mu_1}(\hat{\Sigma}), \dots, \mathrm{col}_{\mu_r}(\hat{\Sigma})$ are linearly independent. Then (\ref{eq:AdmissibilityCond01}) implies that $\lambda_{\mu_1}$, $\dots$, $\lambda_{\mu_r}$ are eigenvalues of $\mathbf{J}$ and $\mathrm{col}_{\mu_1}(\hat{\Sigma}),\dots,\mathrm{col}_{\mu_r}(\hat{\Sigma})$ are the corresponding eigenvectors. If $\hat{\Sigma}$ has an additional nonzero column, $\mathrm{col}_j(\hat{\Sigma})$, with $j\notin\{\mu_1,\dots,\mu_r\}$, then by (\ref{eq:AdmissibilityCond01}) this column is an eigenvector of $\mathbf{J}$ corresponding to the eigenvalue $\lambda_j$ and $\lambda_j\not=\lambda_{\mu_i}$ for $1\leq i\leq r$. On the other hand, $\mathrm{col}_j(\hat{\Sigma})$ is a linear combination of $\mathrm{col}_{\mu_1}(\hat{\Sigma}),\dots,\mathrm{col}_{\mu_r}(\hat{\Sigma})$ which is impossible, since eigenvectors corresponding to different eigenvalues are linearly independent. Thus all columns except $\mathrm{col}_{\mu_1}(\hat{\Sigma}),\dots,\mathrm{col}_{\mu_r}(\hat{\Sigma})$ must be zero.

	Conversely, assume $\hat{\Sigma}$ has $r$ nonzero columns $\mathrm{col}_{\mu_1}(\hat{\Sigma}), \dots, \mathrm{col}_{\mu_r}(\hat{\Sigma})$ and all other columns of $\hat{\Sigma}$ are zero. Since $\mathrm{rank}(\hat{\Sigma}) = r$, these columns are linearly independent. Let $Q$ be a $p\times p$ permutation matrix that maps the column $j$ to the column $\mu_j$ for $1\leq j\leq r$. Then 
	$$
		\begin{array}{ccl}
			\hat{\Sigma} & = & [0,\dots,0,\mathrm{col}_{\mu_1}(\hat{\Sigma}),0,\dots,0,\mathrm{col}_{\mu_r}(\hat{\Sigma}),0,\dots,0]   \\
									 & = & [\mathrm{col}_{\mu_1}(\hat{\Sigma}),\dots,\mathrm{col}_{\mu_r}(\hat{\Sigma}),0,\dots,0]Q.
		\end{array}
	$$
	Let $\hat{\Sigma}_0 = [\mathrm{col}_{\mu_1}(\hat{\Sigma}),\dots,\mathrm{col}_{\mu_r}(\hat{\Sigma})]$ and $\hat{\Sigma}_0^*$ be the adjoint (complex conjugate transpose) matrix of $\hat{\Sigma}_0$ and define 
	$$
		\mathbf{J} = \hat{\Sigma}\Lambda Q^T\left[\begin{array}{c}
			(\hat{\Sigma}_0^*\hat{\Sigma}_0)^{-1}\hat{\Sigma}_0^* \\
			O_{(p-r)\times N}
		\end{array}\right].
	$$
	Then
	$$
		\begin{array}{ccl}
			\mathbf{J}\hat{\Sigma} & = & \hat{\Sigma}\Lambda Q^T\left[\begin{array}{c} (\hat{\Sigma}_0^*\hat{\Sigma}_0)^{-1}\hat{\Sigma}_0^* \\ 
			O_{(p-r)\times N} \end{array}\right][\hat{\Sigma}_0,O_{N\times(p-r)}]Q	\\
			              & = & \hat{\Sigma}\Lambda Q^T\left[\begin{array}{cc} I_{r\times r} & O_{r\times (p-r)} \\ O_{(p-r)\times r} & O_{(p-r)\times(p-r)} \end{array}\right]Q	\\
			              & = & \hat{\Sigma}\Lambda\mathrm{diag}(s_1,\dots,s_p)												\\
			              & = & \hat{\Sigma}\Lambda.
		\end{array}
	$$
	where $I_{r\times r}$ is the identity matrix of size $r\times r$, $O_{m\times n}$ is zero matrix of size $m\times n$, and 
	$$
		s_j = \begin{cases} 1\mathrm{, if }j\in\{\mu_1,\dots,\mu_r\} \\ 0\mathrm{, if }j\notin\{\mu_1,\dots,\mu_r\} \end{cases}.
	$$
\end{proof}

\rem A group theoretical interpretation of admissible cycles $\Sigma$ and the associated matrices $\mathbf{J^0}$ and $\mathbf{J}$ can be given as follows:

\noindent (a) We denote by $\mathbb{Z}_p$ the cyclic group of order $p$ defined by addition of integers modulo $p$. Viewed as permutation group, the generator of $\mathbb{Z}_p$, addition by 1 mod $p$, corresponds to the cyclic permutation $\{0,1,\dots,p-1\}\rightarrow\{1,2,\dots,p-1,0\}$, and the matrix $\mathbf{P}$ is an orthogonal representation of this generator in $\mathbb{R}^p$ or $\mathbb{C}^p$ that cyclically permutes row vectors to the left. Accordingly, the matrices $\mathbf{P}^k$, $0\leq k<p$, form a $p$-dimensional representation of $\mathbb{Z}_p$ with $\mathbf{P}^p = I$, the $p\times p$ identity matrix.

\noindent (b) The admissibility condition $\mathrm{span}\{\mathrm{R}(\Sigma)\}=\mathrm{span}\{\mathrm{R}(\Sigma\mathbf{P})\}$ means that the rows of $\Sigma$ span a subspace of $\mathbb{R}^p$ or $\mathbb{C}^p$ that is invariant under $\mathbf{P}$, and hence under the full representation of $\mathbb{Z}_p$, that is,
	$$
		\mathrm{span}\{\mathrm{R}(\Sigma)\} = \mathrm{span}\{\mathrm{R}(\Sigma\mathbf{P}^k)\}\mbox{, }0\leq k<p.
	$$
	Moreover, with $\mathbf{J} = \Sigma\mathbf{P}\Sigma^+$ and $\Sigma\mathbf{P}\Sigma^+\Sigma = \Sigma\mathbf{P}$, we find that $\mathbf{J}^2 = \Sigma\mathbf{P}^2\Sigma^+$ and inductively
	\begin{equation}
		\label{eq:JPK00}
		\mathbf{J}^k = \Sigma\mathbf{P}^k\Sigma^+\mbox{, }0\leq k<p,
	\end{equation}
	which shows that $\mathbf{J}^p = \Sigma\Sigma^+ = \mathbf{J^0}$ and $\mathbf{J}^k\Sigma = \Sigma\mathbf{P}^k$ for $0\leq k<p$. This means that $\mathbf{J}$ restricted to the column space $\mathrm{span}\{\mathrm{C}(\Sigma)\}$, where $\mathrm{C}(\Sigma)$ denotes the set of column vectors of $\Sigma$, generates a representation of $\mathbb{Z}_p$ in this subspace of $\mathbb{R}^N$. We also note that $\mathbf{J}^0$ is the orthogonal projection onto $\mathrm{span}\{\mathrm{C}(\Sigma)\}$, and 
	\begin{equation}
		\label{eq:J0JK00}
		\mathbf{J^0}\mathbf{J}^k = \mathbf{J}^k\mathbf{J^0} = \mathbf{J}^k,
	\end{equation}
	for all $0\leq k<p$, which is a straightforward consequence of the basic properties $\Sigma\Sigma^+\Sigma = \Sigma$ and $\Sigma^+\Sigma\Sigma^+ = \Sigma^+$ of the pseudoinverse and (\ref{eq:JPK00}). Clearly, the ranks of $\mathbf{J}$, $\mathbf{J^0}$ and $\Sigma$ coincide and are equal to the dimensions of the vector spaces $\mathrm{span}\{\mathrm{C}(\Sigma)\}$ and $\mathrm{span}\{\mathrm{R}(\Sigma)\}$ in which $\mathbb{Z}_p$ acts with matrix generators $\mathbf{J}$ and $\mathbf{P}$, respectively.

\noindent (c) The group $\mathbb{Z}_p$ has exactly $p$ irreducible complex representations which are all one-dimensional and are generated by multiplication of a complex number by $\bar{\rho}^k$, $0\leq k<p$ \citep{SymGroup}. When restricted to real spaces and $\rho^k\notin\mathbb{R}$, the multiplications by $\bar{\rho}^k$ and $\rho^k = \bar{\rho}^{p-k}$ can be combined to form a two-dimensional real irreducible representation space, in which the generator of $\mathbb{Z}_p$ acts by rotation of vectors by the angle $2\pi k/p$. For the representation of $\mathbb{Z}_p$ in the full space of $p$-dimensional row vectors generated by $\mathbf{P}$, the rows in $V^*$ are (complex) basis vectors for these irreducible subspaces, and those basis vectors with eigenvalues $\bar{\rho}^k$ for which $\hat{\Sigma}$ has a nonzero column span the irreducible subspaces in $\mathrm{span}\{\mathrm{R}(\Sigma)\}$. (Real bases in case of $\bar{\rho}^k\notin\mathbb{R}$ are obtained by taking real and imaginary parts of these vectors, but we prefer to use the complex basis vectors.) Likewise, the non-zero columns of $\hat{\Sigma}$ form complex bases of the irreducible subspaces of the $\mathbb{Z}_p$-representation generated by $\mathbf{J}$ in $\mathrm{span}\{\mathrm{C}(\Sigma)\}$. We note that, given a row-vector $x\in\mathbb{C}^p$, $xV$ is the discrete Fourier transform of $x$, and the components of $xV$ are the expansion coefficients of $x$ represented by the basis vectors in $V^*/p$.

Closely related to the group $\mathbb{Z}_p$ are the $p$-th roots of unity, which in turn are intimately related to the cyclotomic polynomials. In Sections 3-5 we will make some use of these polynomials and therefore summarize their basic properties in the appendix. 

\noindent (d) Given a row-vector $x\in\mathbb{R}^p$, the orbit of $x$ under $\mathbf{P}$ is defined as the the set $\{x, x\mathbf{P}, \dots, x\mathbf{P}^{p-1}\}$. For generic $x\in\mathbb{R}^p$, this set is a basis of $\mathbb{R}^p$, however the set of binary row vectors, $x=\eta$, is finite and the orbit of $\eta$, which we call a loop, may span only a proper subspace of $\mathbb{R}^p$. In the next section we classify (admissible) cycles according to the decomposition of $\mathrm{R}(\Sigma)$ into sets of rows belonging to different loops. To pursue this, we will introduce a concept of irreducibility that differs from the standard group-theoretical version above.

\section{Classification of Cycles}
\subsection{Simple Cycles}
\defn Let $\eta = (\eta^1,\eta^2,\dots,\eta^p)$ be a $p$-dimensional binary row vector. The set
$$
	\{\eta_{\nu}:\eta_{\nu} = \eta\mathbf{P}^{\nu}, \nu = 0,1,2,\dots,p-1\},
$$
is called a \emph{loop} and is denoted by $\mathcal{L}_{\eta}$. 

\rem For any loop $\mathcal{L}_{\eta}$, $|\mathcal{L}_{\eta}|\leq p$. More precisely, $|\mathcal{L}_{\eta}| = m$, where $m$ is a factor of $p$. In particular, if $\eta = (+,+,\dots,+)$, then $|\mathcal{L}_{\eta}| = 1$, as $\eta\mathbf{P} = \eta$.

\defn A cycle $\Sigma$ is called \emph{simple}, if its row vectors are from a loop generated by some row vector $\eta$, i.e.,
\begin{equation}
	\label{eq:SimpleCycles00}
	\mathrm{R}(\Sigma)\subseteq\mathcal{L}_{\eta}.
\end{equation}
A cycle $\Sigma$ is \emph{composite}, if it is not simple.

\defn Let $\Sigma$ be a cycle. The set $G_{\Sigma}=\{\eta_1,\eta_2,\dots,\eta_q\}$ is said to be a \emph{generator} of $\Sigma$, if
\begin{equation}
	\label{eq:Generator00}
	\mathcal{L}_{\eta_i}\cap\mathcal{L}_{\eta_j}=\emptyset\mbox{, }\forall i\not=j,
\end{equation}
and
\begin{equation}
	\label{eq:Generator01}
	G_{\Sigma}=\{\eta_1,\eta_2,\dots,\eta_q\}\subseteq \mathrm{R}(\Sigma)\subseteq\bigcup\limits_{i=1}^q\mathcal{L}_{\eta_i}.
\end{equation}

Note that any vector in $\mathcal{L}_{\eta_i} \cap \mathrm{R}(\Sigma)$ can be chosen as generator instead of $\eta_i$ in $G_{\Sigma}$, that is, the generators are unique up to cyclic permutations and the condition to be vectors in $\mathrm{R}(\Sigma)$. In particular, for simple cycles there is only one generator, $|G_{\Sigma}| = 1$, and every row in $\mathrm{R}(\Sigma)$ can be chosen for this generator. A simple criterion for admissibility is the following.

\prop A cycle $\Sigma$ is admissible if $\mathcal{L}_{\eta}\subset\mathrm{span}\{\mathrm{R}(\Sigma)\cap\mathcal{L}_{\eta}\}$ for every $\eta\in G_{\Sigma}$.
\begin{proof}
	This follows immediately from the fact that, under the given hypothesis, every row in $\mathrm{R}(\Sigma\mathbf{P})$ can be represented as linear combination of a subset of rows in $\mathrm{R}(\Sigma)$.
\end{proof}

\defn Let $\eta$ be any row vector. The \emph{rank} of $\eta$ is defined as the dimension of the vector space spanned by the row vectors in the loop generated by $\eta$, i.e.,
\begin{equation}
	\label{eq:DefRVRank00}
	\mathrm{rank}(\eta) = \dim\mathrm{span}\{\mathcal{L}_{\eta}\}.
\end{equation}
	
\thm Let $\Sigma$ be a simple cycle generated by $\eta$, i.e., $\eta\in \mathrm{R}(\Sigma)\subseteq\mathcal{L}_{\eta}$. Then $\Sigma$ is admissible, if and only if
		\begin{equation}
			\label{eq:SimpleAdCondition00}
			\mathrm{rank}(\Sigma) = \mathrm{rank}(\eta).
		\end{equation}
\begin{proof}
	Suppose $\mathrm{rank}(\Sigma) = \mathrm{rank}(\eta)$. Then $\mathrm{span}\{\mathrm{R}(\Sigma)\} = \mathrm{span}\{\mathcal{L}_{\eta}\}$ as $\mathrm{R}(\Sigma)\subseteq\mathcal{L}_{\eta}$. Since $\mathbf{P}$ is nonsingular and $\mathcal{L}_{\eta}\mathbf{P} = \mathcal{L}_{\eta}$, it follows that $\mathrm{span}\{\mathrm{R}(\Sigma\mathbf{P})\} = \mathrm{span}\{\mathcal{L}_{\eta}\}$, hence $\Sigma$ is admissible. Conversely, suppose that $\Sigma$ is admissible, i.e. $\mathrm{span}\{\mathrm{R}(\Sigma)\} = \mathrm{span}\{\mathrm{R}(\Sigma\mathbf{P}^m)\}$ for all $m\in\mathbb{N}$. Assume $\mathrm{rank}(\Sigma) < \mathrm{rank}(\eta)$. Then there exists $\hat{\eta}\in\mathcal{L}_{\eta}$ with $\hat{\eta}\notin\mathrm{span}\{\mathrm{R}(\Sigma)\}$. Let $\hat{\eta} = \eta\mathbf{P}^{\mu}$ for some $0<\mu<p$. Since $\eta \in \mathrm{R}(\Sigma)$, it follows that
	$$
		\hat{\eta}\in\mathrm{R}(\Sigma\mathbf{P}^{\mu})\subset\mathrm{span}\{\mathrm{R}(\Sigma\mathbf{P}^{\mu})\} = \mathrm{span}\{\mathrm{R}(\Sigma)\},
	$$
	which contradicts $\hat{\eta}\notin\mathrm{span}\{\mathrm{R}(\Sigma)\}$.
\end{proof}

\rem In general, although $\mathbf{P}$ preserves the rank of any cycle $\Sigma$, the vector space spanned by the row vectors of $\Sigma$ may not be invariant under $\mathbf{P}$. For simple cycles, the condition (\ref{eq:SimpleAdCondition00}) guarantees that the vector space spanned by the rows of $\Sigma$ is invariant under $\mathbf{P}$, and hence guarantees the admissibility of $\Sigma$. The condition (\ref{eq:SimpleAdCondition00}) will be referred to as \emph{admissibility condition} for simple cycles. 

\subsection{Separable Composite Cycles}
In order to generalize the class of simple admissible cycles to the class of separable composite cycles, we first introduce the concept of decomposability of the row space of a cycle into irreducible subspaces.

\defn Let $\Sigma$ be an admissible cycle of period $p$, and let $\mathcal{U} = \mathrm{span}\{\mathrm{R}(\Sigma)\}$. Note that admissibility implies $\mathcal{U}\mathbf{P} = \mathcal{U}$, i.e. $\mathcal{U}$ is invariant under $\mathbf{P}$. Let $\mathcal{V} \subseteq \mathcal{U}$ be a subspace of $\mathcal{U}$ and assume $\mathcal{V}\mathbf{P} = \mathcal{V}$ and $\mathrm{R}(\Sigma) \cap \mathcal{V} \not= \emptyset$. Then

\noindent (a) The subspace $\mathcal{V}$ is called \emph{reducible} if there exists a proper subspace $\mathcal{W}$ of $\mathcal{V}$ such that $\mathcal{W}\mathbf{P} = \mathcal{W}$ and $\mathrm{R}(\Sigma)\cap\mathcal{W}\not=\emptyset$.

\noindent (b) The subspace $\mathcal{V}$ is said to be \emph{decomposable}, if $\mathcal{V}$ has the direct sum decomposition
		\begin{equation}
			\label{eq:DirectSumDecomp00}
			\mathcal{V} = \mathcal{V}_1\oplus\mathcal{V}_2\oplus\cdots\oplus\mathcal{V}_n,
		\end{equation}
		where $n\geq 2$, $\mathcal{V}_i$ is invariant under $\mathbf{P}$ and $\mathcal{V}_i\cap\mathrm{R}(\Sigma) \not= \emptyset$ for every $1\leq i\leq n$. The subspace $\mathcal{V}$ is said to be \emph{indecomposable}, if it is not decomposable. If $\mathcal{V}_i$ in (\ref{eq:DirectSumDecomp00}) is indecomposable for every $i$, then (\ref{eq:DirectSumDecomp00}) is called a \emph{complete decomposition} of $\mathcal{V}$.

\noindent (c) The vector space $\mathcal{U}$ is called \emph{semisimple}, if $\mathcal{U}$ is the direct sum of irreducible subspaces in the sense of (a). Note that semisimplicity of $\mathcal{U}$ includes the case where $\mathcal{U}$ is irreducible, in which case we call $\mathcal{U}$ simple.

\rem We emphasize that, because our purpose is to study the structure of the invariant subspaces spanned by the row vectors of $\Sigma$, the concepts of \emph{reducibility} and \emph{decomposability} introduced in Definition 6 are slightly different from the standard definitions used in the representation theory of finite groups (we require that each subspace contains a row vector of $\Sigma$). An irreducible/indecomposable invariant subspace in the sense of Definition 6 may be reducible/decomposable in terms of the standard definitions of representation theory applied to the cyclic group $\mathbb{Z}_p$ generated by $\mathbf{P}$.

	It is clear that if $\Sigma$ is simple and admissible, then $\mathcal{U}$ is simple and consequently indecomposable, as $\eta\in\mathcal{U}_i$ implies that $\eta\mathbf{P}^k\in\mathcal{U}_i$ for every $k\in\mathbb{N}$. However, the converse is not necessarily true. In the next example, we show that the vector space spanned by the row vectors of a composite cycle may be reducible but not decomposable.

\ex Consider
$$
	\Sigma = \left(\begin{array}{rrrrrr}
		 + &  + &  + & - &  - &  - \\
		 + &  + &  - & - &  - &  + \\
		 + &  - &  - & - &  + &  + \\
		 + &  - &  + & - &  + &  -
	\end{array}\right),
$$
and let $\eta_j = \mathrm{row}_j(\Sigma)$. Clearly, $\Sigma$ is a composite cycle, as it is generated by $\{\eta_1,\eta_4\}$. Let $\mathcal{U} = \mathrm{span}\{\mathrm{R}(\Sigma)\}$, $\mathcal{U}_1 = \mathrm{span}\{\mathcal{L}_{\eta_1}\}$ and $\mathcal{U}_2 = \mathrm{span}\{\mathcal{L}_{\eta_4}\}$. Since $\eta_4 = \eta_1 - \eta_2 + \eta_3$, i.e. $\eta_4\in\mathcal{U}_1$, we have that  $\mathcal{U}_2\subset\mathcal{U}_1 = \mathcal{U}$. Moreover, $\mathrm{R}(\Sigma\mathbf{P}) = \{\eta_2,\eta_3,-\eta_1,-\eta_4\}$ implies $\mathcal{U}\mathbf{P} = \mathcal{U}$, hence $\Sigma$ is admissible. Since both $\mathcal{U}_1$ and $\mathcal{U}_2$ are invariant under $\mathbf{P}$ it follows that $\mathcal{U}$ is reducible, however, $\mathcal{U}$ is not decomposable.

\prop Let $\Sigma$ be an admissible cycle with generator $G_{\Sigma} = \{\eta_1,\dots,\eta_q\}$. Assume $\mathcal{U} = \mathrm{span}\{\mathrm{R}(\Sigma)\}$ is semisimple and let $\mathcal{U} = \mathcal{U}_1 \oplus \mathcal{U}_2 \oplus \cdots \oplus \mathcal{U}_n$ be a decomposition of $\mathcal{U}$ into irreducible subspaces. Then $n\leq q$ and there exists a subset $\{i_1, \dots, i_n\}\subseteq \{1, \dots, q\}$ such that $\mathcal{U}_j = \mathrm{span}\{\mathcal{L}_{\eta_{i_j}}\}$ for $1\leq j\leq n$. Moreover, if $\mathrm{span}\{\mathcal{L}_{\eta_i}\} \not= \mathrm{span}\{\mathcal{L}_{\eta_j}\}$ for every $i,j\in\{1, \dots, q\}$ with $i\not=j$, then $n=q$.
\begin{proof}
	Let $\eta\in G_{\Sigma}$ and let $\mathcal{V}\in\{\mathcal{U}_j|1\leq j\leq n\}$ be the subspace in the decomposition of $\mathcal{U}$ that contains $\eta$. Invariance of $\mathcal{V}$ implies $\eta\mathbf{P}\in\mathcal{V}$, hence $\eta\mathbf{P}^2 \in \mathcal{V}$ and by induction $\mathcal{L}_{\eta}\subset\mathcal{V}$, thus $\mathrm{span}\{\mathcal{L}_{\eta}\}\subset\mathcal{V}$. Since $\mathrm{span}\{\mathcal{L}_{\eta}\}$ is invariant, $\mathrm{span}\{\mathcal{L}_{\eta}\} \cap \mathrm{R}(\Sigma) \not= \emptyset$ and $\mathcal{V}$ is irreducible, it follows that $\mathcal{V} = \mathrm{span}\{\mathcal{L}_{\eta}\}$, and there exists no $\eta'\in G_{\Sigma}$, $\eta' \not= \eta$, such that $\mathrm{span}\{\mathcal{L}_{\eta'}\}$ is a proper subspace of $\mathrm{span}\{\mathcal{L}_{\eta}\}$ and vice versa. Thus, for $i\not=j$, either $\mathrm{span}\{\mathcal{L}_{\eta_i}\} \cap \mathrm{span}\{\mathcal{L}_{\eta_j}\} = \{\mathbf{0}\}$ or $\mathrm{span}\{\mathcal{L}_{\eta_i}\} = \mathrm{span}\{\mathcal{L}_{\eta_j}\}$. It follows that there exists $i_1,\dots,i_n\in\{1,\dots,q\}$, $i_j\not=i_k$ if $j\not=k$, such that 
	$$
		\mathrm{span}\{\mathrm{R}(\Sigma)\} = \mathrm{span}\{\bigcup\limits_{i=1}^q\mathcal{L}_{\eta_i}\} = \bigoplus\limits_{j=1}^n\mathrm{span}\{\mathcal{L}_{\eta_{i_j}}\}.
	$$
\end{proof}

\ex Let $\eta_1 = (+,+,-,+,+,-)$, $\eta_2 = -\eta_1$ and $\eta_3 = (+,+,+,-,-,-)$. Let $\Sigma$ be the $9\times 6$-cycle defined by
$$
	\Sigma = (\Sigma_1^T,\Sigma_2^T,\Sigma_3^T)^T,
$$
where $\Sigma_j = (\eta_j^T, (\eta_j\mathbf{P})^T,(\eta_j\mathbf{P}^2)^T)^T$ for $j=1,2,3$. Then $G_{\Sigma} = \{\eta_1, \eta_2, \eta_3\}$, $\mathrm{span}\{\mathrm{R}(\Sigma)\} = \mathrm{span}\{\mathcal{L}_{\eta_1}\}\oplus\mathrm{span}\{\mathcal{L}_{\eta_3}\}$, and $\mathrm{span}\{\mathcal{L}_{\eta_2}\}=\mathrm{span}\{\mathcal{L}_{\eta_1}\}$. Clearly, $\mathrm{span}\{\mathcal{L}_{\eta_1}\}$ and $\mathrm{span}\{\mathcal{L}_{\eta_3}\}$ are irreducible, thus $\mathrm{span}\{\mathrm{R}(\Sigma)\}$ is semisimple. Likewise, for the cycle $\Sigma = (\Sigma_1^T,\Sigma_2^T)$, $\mathrm{span}\{\mathrm{R}(\Sigma)\} = \mathrm{span}\{\mathcal{L}_{\eta_i}\}$, $i = 1, 2$, hence $\mathrm{span}\{\mathrm{R}(\Sigma)\}$ is simple.

In general, the vector space $\mathcal{U} = \mathrm{span}\{\mathrm{R}(\Sigma)\}$ of an arbitrary composite cycle $\Sigma$ may have subspaces which are not invariant or do not contain any binary row vector of $\Sigma$, or both. By contrast, if $\mathcal{U}$ is semisimple, $\mathcal{U}$ can be decomposed into irreducible subspaces corresponding to the loops of their generators, but some of these subspaces may coincide. This coincidence is still considered as a degeneracy (see Section 4.3), which we exclude in the class of separable cycles introduced next.

\defn Let $\Sigma$ be a composite cycle with generator $G_{\Sigma}$, i.e. $|G_{\Sigma}|\geq 2$. We call $\Sigma$ \emph{separable}, if $\mathrm{span}\{\mathrm{R}(\Sigma)\}$ is semisimple and $\mathrm{span}\{\mathcal{L}_{\eta}\} \not= \mathrm{span}\{\mathcal{L}_{\eta'}\}$ for any $\eta, \eta' \in G_{\eta}$ with $\eta \not= \eta'$. If $\Sigma$ is not separable, $\Sigma$ is said to be \emph{inseparable}.

Note that the hypotheses for a cycle $\Sigma$ to be separable require that $\mathcal{U} = \mathrm{span}\{\mathrm{R}(\Sigma)\}$ is invariant under $\mathbf{P}$, i.e. separable cycles are a priori admissible.

\thm (\textbf{Separability Condition for Composite Cycles}) Let $\Sigma$ be a composite cycle with generator $G_{\Sigma} = \{\eta_1, \eta_2,\dots,\eta_q\}$. Then $\Sigma$ is separable, if and only if
\begin{equation}
	\label{eq:CompositeAdCondition00}
	\mathrm{rank}(\Sigma) = \sum\limits_{i=1}^q\mathrm{rank}(\eta_i).
\end{equation} 
\begin{proof}
	If $\Sigma$ is separable, (\ref{eq:CompositeAdCondition00}) follows directly from Proposition 3 and Definition 7.
	
	Conversely, suppose (\ref{eq:CompositeAdCondition00}) holds. Since
	$$
		\mathrm{span}\{\mathrm{R}(\Sigma)\} = \mathrm{span}\{(\bigcup\limits_{i=1}^q\mathcal{L}_{\eta_i}) \cap \mathrm{R}(\Sigma)\} \subseteq \mathrm{span}\{\bigcup\limits_{i=1}^q\mathcal{L}_{\eta_i}\},
	$$
	(\ref{eq:CompositeAdCondition00}) implies that $\mathrm{span}\{\mathcal{L}_{\eta_i} \cap \mathrm{R}(\Sigma)\} = \mathrm{span}\{\mathcal{L}_{\eta_i}\}$ for each $1\leq i\leq q$, $\mathrm{span}\{\mathcal{L}_{\eta_i}\} \cap \mathrm{span}\{\mathcal{L}_{\eta_j}\} = \{\mathbf{0}\}$, if $i\not=j$, and hence
	$$
		\mathrm{span}\{\mathrm{R}(\Sigma)\} = \bigoplus\limits_{i=1}^q\mathrm{span}\{\mathcal{L}_{\eta_i}\}.
	$$
	It follows that each $\mathrm{span}\{\mathcal{L}_{\eta_i}\}$ is irreducible, thus $\Sigma$ is separable according to Definitions 6 and 7.
\end{proof}

\ex Consider 
$$
	\Sigma = \left(\begin{array}{rrrrrrrr}
		 + &  + &  + &  + &  - &  - &  - &  - \\
		 + &  + &  + &  - &  - &  - &  - &  + \\
		 + &  + &  - &  - &  - &  - &  + &  + \\
		 + &  - &  - &  - &  - &  + &  + &  + \\
		 + &  + &  - &  - &  + &  + &  - &  - \\
		 + &  - &  - &  + &  + &  - &  - &  + \\
		 + &  - &  + &  - &  + &  - &  + &  -
	\end{array}\right).
$$
We have that $G_{\Sigma}=\{\eta_1,\eta_5,\eta_7\}$, where $\eta_j = \mathrm{row}_j(\Sigma)$, $1\leq j\leq 7$. It is easy to see that $\mathcal{U}_1 = \mathrm{span}\{\mathcal{L}_{\eta_1}\}$, $\mathcal{U}_2 = \mathrm{span}\{\mathcal{L}_{\eta_5}\}$, and $\mathcal{U}_3 = \mathrm{span}\{\mathcal{L}_{\eta_7}\}$ intersect trivially, hence
$$
	\mathcal{U} = \mathrm{span}(\mathrm{R}(\Sigma)) = \mathcal{U}_1\oplus\mathcal{U}_2\oplus\mathcal{U}_3,
$$
which implies that $\Sigma$ is separable, and hence admissible.

\subsection{Inseparable Composite Cycles}
By Definition 7, inseparability of a composite cycle $\Sigma$ happens in two different cases. In the first case, the vector space $\mathcal{U} = \mathrm{span}\{\mathrm{R}(\Sigma)\}$ has a reducible but indecomposable invariant subspace, which entirely contains another invariant subspace as a subspace (see Example 1). This includes the case where $\mathcal{U}$ is semisimple and $\mathrm{span}\{\mathcal{L}_{\eta}\} = \mathrm{span}\{\mathcal{L}_{\eta'}\}$ for two different generators $\eta$ and $\eta'$ (see Example 2). In the second case, the vector space $\mathcal{U}$ has two or more indecomposable (reducible or not) invariant subspaces sharing a nontrivial intersection as common proper subspaces. We now discuss the admissibility of these two types of inseparable composite cycles.

\defn Let $G_{\Sigma}=\{\eta_1,\eta_2,\dots,\eta_q\}$ be a generator of a cycle $\Sigma$. A subset $EG_{\Sigma} = \{\epsilon_1, \epsilon_2,\dots,\epsilon_r\}$, $r\leq q$, of $G_{\Sigma}$ is called an \emph{essential generator} of $\Sigma$, if $EG_{\Sigma}$ is minimal in the sense that

\noindent (a) $\mathrm{span}\{\mathrm{R}(\Sigma)\} \subseteq \mathrm{span}\{\bigcup\limits_{i=1}^r\mathcal{L}_{\epsilon_i}\}$;

\noindent (b) for any $\epsilon_i$, $\epsilon_j\in EG_{\Sigma}$ with $i\not=j$, $\mathrm{span}(\mathcal{L}_{\epsilon_i}) \cap \mathrm{span}(\mathcal{L}_{\epsilon_j})$ is a proper subspace of both $\mathrm{span}(\mathcal{L}_{\epsilon_i})$ and $\mathrm{span}(\mathcal{L}_{\epsilon_j})$;

\noindent (c) for every $\eta_i\in G_{\Sigma}$, if $\eta_i\notin\mathrm{span}\{\bigcup\limits_{j=1,j\not=i}^q\mathcal{L}_{\eta_j}\}$, then $\eta_i\in EG_{\Sigma}$.

Note that an admissible cycle may have different sets of essential generators, i.e. $EG_{\Sigma}$ is in general not unique. Proposition 2 can be directly rephrased in terms of essential generators.

\prop A cycle $\Sigma$ is admissible, if for any essential generator $EG_{\Sigma}$, $\mathcal{L}_{\epsilon} \subset \mathrm{span}\{\mathrm{R}(\Sigma)\cap\mathcal{L}_{\epsilon}\}$ for every $\epsilon\in EG_{\Sigma}$. Conversely, if $\Sigma$ is admissible and $EG_{\Sigma} = \{\epsilon_1,\dots,\epsilon_r\}$ is an essential generator of $\Sigma$, then 
\begin{equation}
	\label{eq:InsepAdCond00}
	\mathrm{rank}(\Sigma) \leq \sum\limits_{i=1}^r\mathrm{rank}(\epsilon_i) \leq |\mathrm{R}(\Sigma)| \leq N.
\end{equation}

\rem The condition (b) in Definition 8 includes three cases. 

\noindent (a) For every $i\not=j$, $\mathrm{span}\{\mathcal{L}_{\epsilon_i}\} \cap \mathrm{span}\{\mathcal{L}_{\epsilon_j}\} = \{\mathbf{0}\}$, and $q=r$. Composite cycles in this case are separable.

\noindent (b) For every $i\not=j$, $\mathrm{span}\{\mathcal{L}_{\epsilon_i}\} \cap \mathrm{span}\{\mathcal{L}_{\epsilon_j}\} = \{\mathbf{0}\}$, but $q<r$. Composite cycles in this case are inseparable and degenerate. In this case $\mathcal{U}$ may be semisimple or not, and has the complete decomposition $\mathcal{U} = \bigoplus\limits_{i=1}^q \mathrm{span}\{\mathcal{L}_{\epsilon_i}\}$ (see Proposition 3). Accordingly, we have that a degenerately inseparable composite cycle $\Sigma$ is admissible, if and only if $\mathrm{rank}(\Sigma) = \sum\limits_{i=1}^q\mathrm{rank}(\epsilon_i)$. This generalizes separable cycles and includes, for example, the case where for some $\eta\in\mathrm{R}(\Sigma)$ also $-\eta\in\mathrm{R}(\Sigma)$, but $\mathcal{L}_{\eta} \cap \mathcal{L}_{-\eta} = \emptyset$ (see Example 2).

\noindent (c) For some $i\not=j$, $\mathrm{span}\{\mathcal{L}_{\epsilon_i}\} \cap \mathrm{span}\{\mathcal{L}_{\epsilon_j}\}$ is a nontrivial proper subspace of both invariant subspaces $\mathrm{span}\{\mathcal{L}_{\epsilon_i}\}$ and $\mathrm{span}\{\mathcal{L}_{\epsilon_j}\}$. Composite cycles in this case are genuinely inseparable. This type of cycles is more complicated than the other two. We next study the structure of this type of cycles, and establish an admissibility condition.

\prop Let $\eta$ and $\tilde{\eta}$ be two $p$-dimensional row vectors. If $\eta\in\mathrm{span}\{\mathcal{L}_{\tilde{\eta}}\}$, then $\mathrm{span}\{\mathcal{L}_{\eta}\}\subseteq\mathrm{span}\{\mathcal{L}_{\tilde{\eta}}\}$.
\begin{proof}
	If $\eta\in\mathcal{L}_{\tilde{\eta}}$, we are done. Suppose $\eta\in\mathrm{span}\{\mathcal{L}_{\tilde{\eta}}\}$, i.e., $\eta = \sum\limits_{\nu=1}^p\alpha_{\nu}\tilde{\eta}\mathbf{P}^{\nu}$, $\alpha_{\nu}\in\mathbb{R}$, but $\eta\notin\mathcal{L}_{\tilde{\eta}}$, then for every $t\in\{1,2,\dots,p\}$, $\eta\mathbf{P}^t = \left(\sum\limits_{\nu=1}^p\alpha_{\nu}\tilde{\eta}\mathbf{P}^{\nu}\right)\mathbf{P}^t = \sum\limits_{\nu=1}^{p}\left(\alpha_{\nu}\tilde{\eta}\mathbf{P}^{\nu+t}\right)$, i.e., $\mathcal{L}_{\eta} \subseteq \mathrm{span}\{\mathcal{L}_{\tilde{\eta}}\}$, hence $\mathrm{span}\{\mathcal{L}_{\eta}\}$ $\subseteq$ $\mathrm{span}\{\mathcal{L}_{\tilde{\eta}}\}$.
\end{proof}

\rem Proposition 5 tells that if a row vector is in the vector space spanned by the loop generated by another row vector of the same dimension, then the vector space spanned by the loop generated by this row vector is a subspace of the vector space spanned by the other one. Since for any genuinely inseparable composite cycle, at least two indecomposable invariant subspaces intersect nontrivially, it is natural to ask:

\noindent (a) Does there exist a row vector such that this nontrivial intersection is spanned by the loop generated by it?

\noindent (b) If this row vector exists, can it be $\{-1,1\}$-valued?

	As we will see below in Proposition 6, the answer to the first question is affirmative, however, it remains unclear whether there always exists a binary row vector such that the loop generated by it spans the nontrivial intersection of two indecomposable invariant subspaces. The approach we will use in the proof of Proposition 6 only guarantees the existence of a genuine row vector, which may or may not be binary.
	
Let $V$ be defined as in Theorem 1, i.e. $V = (v^{(0)},v^{(1)},\dots,v^{(p-1)})$, where $v^{(k)} = (1,\rho^k,\rho^{2k},\dots,\rho^{(p-1)k})^T$ and $\displaystyle{\rho = e^{2\pi\mathbf{i}/p}}$.

\defn A row vector $\eta$ (not necessarily binary) is said to \emph{annihilate} the column $v^{(k)}$ of $V$, if $\eta v^{(k)} = 0$, i.e. the two vectors are orthogonal.

Note that, since $v^{(k)}$ is an eigenvector of $\mathbf{P}$ and all eigenvalues of $\mathbf{P}$ are nonzero, $\eta$ annihilates $v^{(k)}$ if and only if $\eta\mathbf{P}^{\nu}$ annihilates $v^{(k)}$ for every $\nu\in\mathbb{Z}$. We need the following fact about the eigenvectors and eigenvalues of circulant matrices, see, e.g., \cite{MatrixTheor}.

\lem Let $\eta = (\eta^1,\dots,\eta^p)$ be an arbitrary real and nonzero row vector, and let $\Sigma_{\eta}$ be the $p\times p$-matrix defined by $\mathrm{row}_j(\Sigma_{\eta}) = \eta(\mathbf{P}^{T})^{(j-1)}$ for $1\leq j\leq p$. Then $V^*\Sigma_{\eta} = \Lambda_{\eta}V^*$, where $\Lambda_{\eta} = \mathrm{diag}(\lambda_{\eta,1},\dots,\lambda_{\eta,p})$ with $\lambda_{\eta,k} = \sum\limits_{j=1}^p\eta^j\rho^{(j-1)(k-1)}$.

\rm Extending Definition 2 to non-binary real row vectors and noting that $\mathrm{R}(\Sigma_{\eta}) = \mathcal{L}_{\eta}$, an immediate consequence of Lemma 1 is the following:

\cor Assume that $\eta v^{(j)}\not=0$ if and only if $j\in\{k_1,\dots,k_s\}\subset\mathbb{Z}_p$, where $\mathbb{Z}_p = \{0, 1, 2, \dots, p-1\}$. Then $\{v^{(k_1)*},\dots,v^{(k_s)*}\}$ is a (complex) basis for $\mathrm{span}\{\mathcal{L}_{\eta}\}$.

\prop Let $\Sigma$ be a cycle with essential generator $EG_{\Sigma}=\{\epsilon_1, \dots, \epsilon_r\}$. Assume that for some $i\not=j$ the indecomposable subspaces $\mathcal{U}_i = \mathrm{span}(\mathcal{L}_{\epsilon_i})$ and $\mathcal{U}_j = \mathrm{span}(\mathcal{L}_{\epsilon_j})$ intersect nontrivially, and $\mathcal{U}_i\cap\mathcal{U}_j$ is a proper subspace of both $\mathcal{U}_i$ and $\mathcal{U}_j$. Then there exists a row vector $\eta$ such that $\mathcal{U}_i\cap\mathcal{U}_j = \mathrm{span}\{\mathcal{L}_{\eta}\}$.
\begin{proof}
	Assume that $\epsilon_i v^{(k)}\not=0$ and $\epsilon_j v^{(k)}\not=0$ if and only if $k\in K_i\subset\mathbb{Z}_p$ and $k\in K_j\subset\mathbb{Z}_p$, respectively. Assume further that $K \equiv K_i\cap K_j = \{k_1,\dots,k_s\}$. According to Corollary 1, $\{v^{(k_1)*},\dots, v^{(k_s)*}\}$ is a basis for $\mathcal{U}_i \cap \mathcal{U}_j$. Let $p_i(x)$ and $p_j(x)$, $x\in\mathbb{C}$, be the polynomials $p_i(x) = \epsilon_i\mathbf{x}$, $p_j(x) = \epsilon_j\mathbf{x}$, where $\mathbf{x} = (1,x,x^2,\dots,x^{p-1})^T$. Since the row vectors defined by the coefficients of $p_i(x)$ and $p_j(x)$ annihilate exactly the $v^{(k)}$ with $k\in\mathbb{Z}_p\backslash K_i$ and $k\in\mathbb{Z}_p\backslash K_j$, respectively, $p_i(x)$ and $p_j(x)$ contain the minimal polynomials of $\rho^k$ for every $k\in\mathbb{Z}_p\backslash K_i$ and $k\in\mathbb{Z}_p\backslash K_j$, respectively, as factors. Multiplying these factors yields a polynomial $p_{ij}(x)$ of degree $\leq p-1$ with $p_{ij}(\rho^k) = 0$ for every $k\in\mathbb{Z}_p\backslash K$ and $p_{ij}(\rho^k) \not= 0$ for every $k\in K$. Set $p_0(x) = p_{ij}(x)$ if the degree of $p_{ij}(x)$ is $p-1$. If the degree of $p_{ij}(x)$ is $<p-1$, set $p_0(x) = p_{ij}(x)\tilde{p}_{ij}(x)$, where $\tilde{p}_{ij}(x)$ is any polynomial such that $\tilde{p}_{ij}(\rho^k)\not=0$ for every $k\in K$ and the degree of $p_0(x)$ is $p-1$. Let $\eta$ be the row vector of coefficients of $p_0(x)$. Then $\eta$ and $\eta\mathbf{P}^{\nu}$ for any $\nu\in\mathbb{Z}$ annihilate every $v^{(k)}$ for $k\in\mathbb{Z}_p\backslash K$, and $\eta v^{(k)}\not=0$ for every $k\in K$, hence $\mathrm{span}\{\mathcal{L}_{\eta}\} = \mathrm{span}\{v^{(k_1)*}, \dots, v^{(k_s)*}\} = \mathcal{U}_i \cap \mathcal{U}_j$.
\end{proof}

\ex In this example, we demonstrate how to find a row vector as claimed in Proposition 6 with the method described in the proof. Consider the composite cycle with $N=10$ and $p=18$ defined by $\Sigma = (\Sigma_1^T, \Sigma_2^T)^T$, where
$$
	\Sigma_1 = (\epsilon_1^T, (\epsilon_1\mathbf{P})^T, \dots, (\epsilon_1\mathbf{P}^6)^T)^T, \Sigma_2 = (\epsilon_2^T, (\epsilon_2\mathbf{P})^T, (\epsilon_1\mathbf{P}^2)^T)^T, 
$$
and 
$$
	\begin{array}{l}
		\epsilon_1 = (\begin{array}{cccccccccccccccccc} 
			+ & + & + & + & + & + & + & - & + & - & - & - & - & - & - & - & + & - 
		\end{array}), \\
		\epsilon_2 = (\begin{array}{cccccccccccccccccc} 
			+ & + & + & - & - & - & + & + & + & - & - & - & + & + & + & - & - & -
		\end{array}).
	\end{array}
$$
This is a genuinely inseparable composite cycle with $G_{\Sigma} = EG_{\Sigma} = \{\epsilon_1,\epsilon_2\}$. The polynomials $p_1(x)$ and $p_2(x)$ can be factorized as follows,
$$
	\begin{array}{l}
		p_1(x) = (1-x)(1+x+x^2)(1-x+x^2)(1+x^3+x^6)(1+2x+2x^2+x^3+x^6), \\
		p_2(x) = (1-x)(1+x+x^2)(1-x^3+x^6)(1+x^3+x^6)(1+x-x^2).
	\end{array}
$$
The factors $1-x$, $1+x+x^2$, $1+x^3+x^6$, $1-x+x^2$ and $1-x^3+x^6$ are cyclotomic factors (see e.g. \cite{DummitFoote}), and the sum of their degrees happens to be 17. Multiplying them out gives
$$
	p_0(x) = \sum\limits_{j=0}^{17}(-1)^j x^j,
$$
thus the row vector $\eta$ constructed in the proof of Proposition 6 is obtained as the binary vector with alternating signs, $\eta = (+,-,+,-,\dots,+,-)$, and $\mathrm{span}\{\mathcal{L}_{\eta}\} = \mathrm{span}\{\eta\}$. One can easily verify that $\mathrm{span}\{\eta\} = \mathrm{span}\{\mathcal{L}_{\epsilon_1}\} \cap \mathrm{span}\{\mathcal{L}_{\epsilon_2}\}$.

Based on their structural features and using a simple inclusion-exclusion argument, an admissibility condition for inseparable composite cycles can be formulated as follows.

\thm (\textbf{Admissibility Condition for Inseparable Composite Cycles}) Let $\Sigma$ be a cycle with essential generator $EG_{\Sigma}=\{\epsilon_1,\dots,\epsilon_r\}$. Then $\Sigma$ is admissible if and only if
\begin{equation}
	\label{eq:CompositeAdCondition01}
	\begin{array}{ccl}
		\mathrm{rank}(\Sigma) & = & \sum\limits_{i=1}^{r}\mathrm{rank}(\epsilon_i) - \sum\limits_{i,j;i\not=j}^{r}\dim(\mathrm{span}\{\mathcal{L}_{\epsilon_i}\} \cap \mathrm{span}\{\mathcal{L}_{\epsilon_j}\})	\\
	                      & + & \sum\limits_{i,j,k;i\not=j\not=k}^{r}\dim(\mathrm{span}\{\mathcal{L}_{\epsilon_i}\} \cap \mathrm{span}\{\mathcal{L}_{\epsilon_j}\} \cap \mathrm{span}\{\mathcal{L}_{\epsilon_k}\})	\\
	                      & - & \cdots \pm \dim(\bigcap\limits_{i=1}^r\mathrm{span}\{\mathcal{L}_{\epsilon_i}\}).
	\end{array}
\end{equation}

\rem The admissibility condition (\ref{eq:CompositeAdCondition01}) is valid for any cycle, and the conditions (\ref{eq:CompositeAdCondition00}) and (\ref{eq:SimpleAdCondition00}) for separability and admissibility of simple cycles, respectively, can be thought of as special cases thereof.

\ex Let $\Sigma$ be the cycle from Example 4. One can easily verify that $\mathrm{rank}(\epsilon_1) = 7$ and $\mathrm{rank}(\epsilon_2) = 3$. Since $\dim(\mathrm{span}\{\mathcal{L}_{\epsilon_1}\} \cap \mathrm{span}\{\mathcal{L}_{\epsilon_2}\}) = 1$, this cycle is admissible.

\section{Network Topology}
To simplify the discussion, we exclude in this section multiple appearances of a binary row vector in a cycle, that is, we consider only cycles $\Sigma$ with $|\mathrm{R}(\Sigma)| = N$.

The classification of cycles $\Sigma$ in Section 4 was based on the decomposition of $\mathrm{R}(\Sigma)$ into subsets of rows associated with disjoint loops. It is, therefore, natural to identify the neurons corresponding to the same loop with a cluster. However, if a cycle has fewer essential generators than generators, the row vectors of a non-essential generator must be combined with one or more essential generators and, moreover, there may be several choices for essential generators. We therefore make the simplifying assumption that all generators are essential generators. For admissible cycles $\Sigma$ this means that for any two distinct generators $\eta_1$, $\eta_2$, the intersection of their spaces $\mathrm{span}\{\mathcal{L}_{\eta_j} \cap \mathrm{R}(\Sigma)\}$, $j = 1, 2$, is a proper subspace of both of them ($\{\mathbf{0}\}$ if the cycle is separable). An immediate consequence of this assumption is that
\begin{equation}
	\label{eq:RankRelat00}
	\mathrm{rank}(\Sigma) \leq \sum\limits_{j=1}^{q}\mathrm{rank}(\eta_j) \leq N,
\end{equation}
if $\Sigma$ is admissible and $G_{\Sigma} = \{\eta_1,\dots,\eta_q\}$. The clusters are isolated if and only if $\Sigma$ is separable. If $\Sigma$ is inseparable, some of the clusters are connected.

Regarding the connectivity within a cluster, linear dependences among its row vectors will prevent any special structure. We call cycles for which such dependences do not occur minimal.

\defn An admissible cycle $\Sigma$ with generator $G_{\Sigma} = \{\eta_1, \eta_2, \dots, \eta_q\}$ is \emph{minimal}, if $EG_{\Sigma} = G_{\Sigma}$ and for every $1\leq i\leq q$,
\begin{equation}
	\label{eq:MinDef00}
	|\mathcal{L}_{\eta_i}\cap\mathrm{R}(\Sigma)| = \mathrm{rank}(\eta_i).
\end{equation}

\rem If $\Sigma$ is a minimal simple or separable composite cycle, then $\Sigma$ is of full row rank. If $\Sigma$ is a minimal inseparable cycle, then the row vectors in $\mathrm{R}(\Sigma) \cap \mathcal{L}_{\eta_i}$ form a basis of $\mathrm{span}\{\mathcal{L}_{\eta_i}\}$ for every $\eta_i\in G_{\Sigma}$. Thus for any minimal cycle $\Sigma$, (\ref{eq:RankRelat00}) holds and the inequalities become equalities if and only if $\Sigma$ is simple or separable. In this case, $\Sigma$ has full row rank and $\Sigma^+ = \Sigma^T(\Sigma\Sigma^T)^{-1}$, which implies that $\mathbf{J^0} = \Sigma\Sigma^+ = I$, the $N\times N$ identity matrix.

For any two $(N\times p)$-cycles $\Sigma$ and $\Sigma'$ with $\mathrm{R}(\Sigma) = \mathrm{R}(\Sigma')$, the cycle matrices are related to each other by $\Sigma' = \mathbf{Q}\Sigma$, where $\mathbf{Q}$ is an $N\times N$ permutation matrix. If in addition $\Sigma$ is admissible with connectivity matrix $\tilde{\mathbf{J}}$, then $\Sigma'$ is also admissible and has connectivity matrix $\tilde{\mathbf{J'}} = \mathbf{Q}\tilde{\mathbf{J}}\mathbf{Q}^{-1}$. Accordingly, if $\mathbf{u}$ is the state of the network with connectivity matrix $\tilde{\mathbf{J}}$, then $\mathbf{u}' = \mathbf{Q}\mathbf{u}$ is the network state corresponding to $\tilde{\mathbf{J}}'$, and solutions $\mathbf{u}(t)$ and $\mathbf{u'}(t)$ of the corresponding differential equations are just permutations of each other as $\tanh(\mathbf{u'}) = \tanh(\mathbf{Q}^{-1}\mathbf{u}) = \mathbf{Q}^{-1}\tanh(\mathbf{u})$.

Without loss of generality, we therefore may assume that a minimal cycle with generators $\eta_1, \dots, \eta_q$ has the form
\begin{equation}
	\label{eq:MinCompCyc}
	\Sigma = (\Sigma_1^T, \Sigma_2^T, \dots, \Sigma_q^T)^T,
\end{equation}
where $\mathrm{R}(\Sigma_j)\subseteq\mathcal{L}_{\eta_j}$, $1\leq j\leq q$, and the vectors in $\Sigma_j$ are sorted from top to bottom as $\eta_j, \eta_{j_1}, \eta_{j_2}, \dots$, with $\eta_{j_i} = s_{j_i}\eta_{j}\mathbf{P}^{\nu_{j_i}}$, $s_{j_i} = 1$ if $-\eta_j\notin\mathcal{L}_{\eta_j}$ and $s_{j_i}\in\{-1,1\}$ if $-\eta_j\in\mathcal{L}_{\eta_j}$, and $0<\nu_{j_i}<\nu_{j_k}$ if $i<k$. We call this form the \emph{standard form} of a minimal cycle.

The minimality requirement does not suffice in general to induce a special network topology within the clusters. We have to require in addition that the powers in the $\Sigma_j$ are consecutive.

\defn A minimal cycle $\Sigma$ in standard form is said to be a \emph{minimal consecutive cycle}, or briefly MC-cycle, if the powers of $\mathbf{P}$ in $\Sigma_j$ above are consecutive, that is, $\nu_{j_i} = i$ for all $1\leq i< \mathrm{rank}(\eta_j)$.

In order that Definition 11 is consistent with the minimality requirement, the rows in $\Sigma_j$ must be linearly independent. The next proposition shows that this is indeed the case, where for simplicity we consider only the case $s_{j_i}=1$.

\prop Let $\eta\not=0$ be any $p$-dimensional row vector with $\mathrm{rank}(\eta) = k$. Then the vectors $\{\eta,\eta\mathbf{P},\eta\mathbf{P}^2,\dots,\eta\mathbf{P}^{k-1}\}$ are linearly independent.
\begin{proof}
	Let $s$ be the smallest positive integer such that $\{\eta,\eta\mathbf{P},\dots,\eta\mathbf{P}^{s-1}\}$ are linearly independent. Then $\eta\mathbf{P}^s$ is a linear combination of $\{\eta,\eta\mathbf{P},\dots,\eta\mathbf{P}^{s-1}\}$,
	\begin{equation}
		\label{eq:MCLD00}
		\eta\mathbf{P}^s = \sum\limits_{\nu=0}^{s-1}\alpha_{\nu}\eta\mathbf{P}^{\nu}.
	\end{equation}
	Right-multiplying this equation by $\mathbf{P}$ yields a representation of $\eta\mathbf{P}^{s+1}$ as linear combination of $\{\eta\mathbf{P},\eta\mathbf{P}^2,\dots,\eta\mathbf{P}^s\}$, and replacing $\eta\mathbf{P}^s$ in this representation by (\ref{eq:MCLD00}) shows that $\eta\mathbf{P}^{s+1}$ is also a linear combination of $\{\eta, \eta\mathbf{P}, \dots, \eta\mathbf{P}^{s-1}\}$. By induction we find that, for any $0\leq \nu < s-p$, $\eta\mathbf{P}^{s+\nu}$ is a linear combination of $\{\eta,\eta\mathbf{P},\dots,\eta\mathbf{P}^{s-1}\}$, hence this set is a basis for $\mathcal{L}_{\eta}$.
\end{proof}

\rm For simple MC-cycles there is only one cluster. In Subsection 5.1 we discuss the possible connectivity structures in such networks in some detail, including the possible values of $N$ for a given $p$, and we also comment on the network topology of simple minimal but non-consecutive cycles. Semisimple MC-cycles consist of isolated clusters corresponding to the different loops in the cycle. Each of these loops forms a simple MC-cycle, and we just give an example in Subsection 5.2. Inseparable minimal (consecutive or non-consecutive) cycles are more complicated and will be discussed in Subsection 5.3. In Subsection 5.4 we demonstrate the effects of fewer essential generators than generators by two examples.

\subsection{Simple MC-Cycles}
\subsubsection{Network Topology}
According to Definition 11, a simple MC-cycle has the form 
\begin{equation}
	\label{eq:StandardForm01}
	\Sigma = (\eta^T,s_1(\eta\mathbf{P})^T,s_2(\eta\mathbf{P}^2)^T,\dots,s_{N-1}(\eta\mathbf{P}^{N-1})^T)^T,
\end{equation}
with $\mathrm{rank}(\eta) = N \leq p$ and $s_i\in\{-1,1\}$. Since the image of the last row vector of $\Sigma$ under $\mathbf{P}$ is a linear combination of the row vectors of $\Sigma$, $\Sigma\mathbf{P}$ has the form $\Sigma\mathbf{P} = \mathbf{A}\Sigma$, where
\begin{equation}
	\label{eq:J1Mini00}
	\mathbf{A} = \left(\begin{array}{cccccc}
		0      & s_1    & 0      & \dots  & 0       				& 0 							\\
		0      & 0      & s_1s_2 & \dots  & 0       				& 0 							\\
		\vdots & \vdots & \vdots & \ddots & \vdots  				& \vdots					\\
		0      & 0      & 0      & \dots  & s_{N-3}s_{N-2}	& 0 							\\
		0      & 0      & 0      & \dots  & 0       				& s_{N-2}s_{N-1}	\\
		a_1    & a_2    & a_3    & \dots  & a_{N-1} 				& a_N
	\end{array}\right),
\end{equation}
with $a_1,\dots,a_{N}\in\mathbb{R}$ and $a_1\not=0$. Moreover, since $\Sigma$ has full row rank, $\Sigma^+ = \Sigma^T(\Sigma\Sigma^T)^{-1}$, which implies
\begin{equation}
	\label{eq:J1Mini01}
	\mathbf{J} = \Sigma\mathbf{P}\Sigma^+ = \mathbf{A}.
\end{equation}
Equations (\ref{eq:J1Mini00}) and (\ref{eq:J1Mini01}) show that the network constructed from a simple MC-cycle consists of a feed-forward chain from the $N$th neuron to the first neuron, and feedback to the $N$th neuron from the subset of the neurons for which $a_i\not=0$, which in any case includes the first neuron. If $a_j = 0$ for $j>1$, then $a_1=\pm 1$, and the network topology is that of a ring, with either excitatory ($a_1 = 1$, $\mathbf{J} = \mathbf{P}^T$ if all $s_j = 1$) or inhibitory connection ($a_1 = -1$) from neuron 1 to neuron $N$. Vectors of the form $\eta = (\sigma,\sigma)$ or $(\sigma,-\sigma)$ have $\mathrm{rank}(\eta)\leq p/2$, and if $\mathrm{rank}(\eta) = p/2 = N$ we have either of these two types of ring structures (see Subsection 5.1.3.).

\ex In Figure 3, A and B, we illustrate the topology of the networks constructed from the following two simple MC-cycles,
$$
	\Sigma = \left(\begin{array}{rrrrrrrrrrrrrr}
		 + &  + &  + &  + &  + &  + &  + & - &  - &  - &  - &  - &  - &  - \\
		 + &  + &  + &  + &  + &  + &  - & - &  - &  - &  - &  - &  - &  + \\
		 + &  + &  + &  + &  + &  - &  - & - &  - &  - &  - &  - &  + &  + \\
		 + &  + &  + &  + &  - &  - &  - & - &  - &  - &  - &  + &  + &  + \\
		 + &  + &  + &  - &  - &  - &  - & - &  - &  - &  + &  + &  + &  + \\
		 + &  + &  - &  - &  - &  - &  - & - &  - &  + &  + &  + &  + &  + \\
		 + &  - &  - &  - &  - &  - &  - & - &  + &  + &  + &  + &  + &  +
	\end{array}\right),
$$
and
$$
	\tilde{\Sigma} = \left(\begin{array}{rrrrrr}
		 + &  + &  - &  + &  - &  - \\
		 + &  - &  + &  - &  - &  + \\
		 - &  + &  - &  - &  + &  + \\
		 + &  - &  - &  + &  + &  - \\
		 - &  - &  + &  + &  - &  +
	\end{array}\right),
$$
respectively. The cycle $\Sigma$ has a ``repeating block structure'', $\Sigma = [B,-B]$, where $B$ is the block consisting of the first 7 columns of $\Sigma$ ($N = p/2 = 7$). This causes the image of the last row to be the negative copy of the first row, i.e. $\eta_7\mathbf{P} = -\eta_1$, where $\eta_i = \mathrm{row}_i(\Sigma)$. It follows that $a_1 = -1$ and $a_i=0$ for $i>1$, thus the first neuron only sends an inhibitory feedback to the seventh neuron (Figure 3A). Similarly, for the cycle $\tilde{\Sigma}$, $N=p-1=5$, and in this case the image of the last row is a linear combination of all other row vectors, $\tilde{\eta}_5\mathbf{P} = -\tilde{\eta}_1 - \tilde{\eta}_2 - \tilde{\eta}_3 - \tilde{\eta}_4 - \tilde{\eta}_5$, where $\tilde{\eta}_i = \mathrm{row}_i(\tilde{\Sigma})$. Accordingly for this cycle $a_i=-1$ for every $i$, i.e. every neuron sends inhibitory feedback to the fifth neuron in the network (Figure 3B).

The two examples above demonstrate that the value of $N = \mathrm{rank}(\eta)$ plays an important role for the network topology of simple MC-cycles. We discuss possible values of $N$ for given cycle-lengths $p$ in the next paragraph.

For minimal but non-consecutive cycles with $N < p-1$ we can have ``gaps'' in the standard form which lead to feedforward chains interrupted by neurons with higher connectivities. The next example demonstrates this possibility.

\ex Consider $\eta = (+,+,+,+,+,+,-,-,-)$ $(p=9)$, and 
$$
	\Sigma = (\eta^T, (\eta\mathbf{P})^T, (\eta\mathbf{P}^2)^T, (\eta\mathbf{P}^4)^T, (\eta\mathbf{P}^5)^T, (\eta\mathbf{P}^6)^T, (\eta\mathbf{P}^8)^T)^T.
$$
This cycle is minimal as $\mathrm{rank}(\eta) = \mathrm{rank}(\Sigma) = 7$, but not consecutive. The gaps are between the third and fourth rows, and the sixth and seventh rows. Since the seventh and first rows are consecutive, there are no other gaps. The connectivity matrix is
$$
	\mathbf{J} = \left(\begin{array}{ccccccc}
		 0 &  1 &  0 &  0 &  0 &  0 &  0 \\
		 0 &  0 &  1 &  0 &  0 &  0 &  0 \\
		-1 &  0 &  1 &  0 &  1 & -1 &  1 \\
		 0 &  0 &  0 &  0 &  1 &  0 &  0 \\
		 0 &  0 &  0 &  0 &  0 &  1 &  0 \\
		 0 & -1 &  1 & -1 &  1 &  0 &  1 \\
		 1 &  0 &  0 &  0 &  0 &  0 &  0
	\end{array}\right),
$$
and shows that we still have the forward chain $3\rightarrow 2\rightarrow 1\rightarrow 7\rightarrow 6\rightarrow 5\rightarrow 4$, but neurons 3 and 6 receive multiple inputs. The network topology is shown in Figure 3C.

\subsubsection{$N$-$p$ relations}
\defn Let $N_p:\mathbb{X}^p\rightarrow\Omega$ be the function defined by $N_p(\eta) = n$ if $\eta$ annihilates $(p-n)$ columns of $V$, where $V$ is defined as in Theorem 1, $\mathbb{X}^p$ is the set of binary row vectors of length $p$, and $\Omega = \{1,2,\dots,p-1\}$.

\rem It is a direct consequence of Theorem 1 that $N_p(\eta) = \mathrm{rank}(\eta)$. Therefore, for a given value of $p$, the image-set $N_p(\mathbb{X}^p)$ contains all possible values of $N$ for which there exists $\eta\in\mathbb{X}^p$ such that (\ref{eq:StandardForm01}) defines a simple MC-cycle. Furthermore, in Section 4.3 we have associated with $\eta\in\mathbb{X}^p$ the polynomial $p_{\eta}(x) = \eta(1,x,\dots,x^{p-1})^T$, where $x$ is a complex variable. Since $p_{\eta}(\rho^k) = \eta v^{(k)}$, $\eta v^{(k)} = 0$ (i.e. $\eta$ annihilates $v^{(k)}$) if and only if $p_{\eta}(x)$ has a factor which is a multiple of the minimal polynomial of $\rho^k$. Thus $N_p(\eta) = \mathrm{rank}(\eta)$ is intimately related to the factorization of $p_{\eta}(x)$.

There appears to be no general characterization of or formula for $N_p(\eta)$. Even for row vectors with repeating block structure such as $\eta = (\sigma, -\sigma)$ or $(\sigma,-\sigma,\sigma)$ ($\sigma\in\mathbb{X}^{p/2}$ or $\mathbb{X}^{p/3}$), the factorization of $p_{\eta}(x)$ does not reveal a formalizable pattern. We therefore just list the sets $N_p(\mathbb{X}^p)$ in Table 1 for $1\leq p\leq 20$. Note that $N_q(\mathbb{X}^q)\subset N_p(\mathbb{X}^p)$ if $q$ divides $p$, since $\mathrm{rank}(\eta) = \mathrm{rank}(\sigma)$ if $\eta = (\sigma, \sigma, \dots, \sigma)$ ($p/q$ repetitions) and $\sigma\in\mathbb{X}^q$. We therefore include in Table 1 only those values $N\in N_p(\mathbb{X}^p)$, for which there exists a row vector $\eta\in\mathbb{X}^p$ with $N = \mathrm{rank}(\eta)$, and $\eta$ is NOT a repetition of some shorter vector $\sigma$. To illustrate how Table 1 was obtained, we compute $N_p(\eta)$ for a row vector with $p = 6$ in Example 8.

\rem Vectors $\eta$ of the form $\eta = (\sigma, \sigma, \dots)\in\mathbb{X}^p$ with $\sigma\in \mathbb{X}^q$ have minimal period $\leq q$ under cyclic permutations. The number of binary vectors of minimal period $p$ is found by subtracting the number of all vectors with smaller minimal period from $2^p$. An inclusion/exclusion argument shows that this number is given by
$$
	M_p = 2^p - \sum\limits_{k=1}^s (-1)^{k-1}\sum\limits_{1\leq i_1<\dots<i_k\leq s} 2^{p/(p_{i_1}p_{i_2}\cdots p_{i_k})},
$$
if $p_1$, $p_2$, $\dots$, $p_s$ are the distinct prime numbers occurring in the prime factorization of $p$ ($M_p = 2^p - 2$ if $p$ is prime). Accordingly, the number of maximal loops, i.e. loops with $|\mathcal{L}_{\eta}| = p$, is $M_p/p$.

\begin{quote}
	\begin{table}
		\caption{Values of $N\in N_p(\mathbb{X}^p)$ attained by some $\eta\in\mathbb{X}^p$ that is NOT of the form $(\sigma, \sigma, \dots, \sigma)$ with $\sigma\in\mathbb{X}^q$, $q<p$, for $p\leq 20$.}
		\begin{center}
			\begin{tabular}{||c|c||c|c||}
				\hline\hline
					$p$ & $N$ 					& $p$ & $N$																	\\
				\hline\hline
					 1	& 1							& 11	&	11																	\\
					 2	&	1							& 12	&	6,7,8,9,10,11,12										\\
					 3	&	3							& 13	&	13																	\\
					 4	&	2,4						& 14	&	7,13,14															\\
					 5	&	5							& 15	&	11,13,15														\\
					 6	&	3,5,6					&	16	&	8,10,11,12,13,14,15,16							\\
					 7	&	7							&	17	&	17																	\\
					 8	&	4,6,7,8				&	18	&	7,9,11,12,13,14,15,16,17,18					\\
					 9	&	7,9						&	19	&	19																	\\
					10	& 5,9,10				&	20	&	10,12,13,14,15,16,17,18,19,20				\\
			\hline\hline
		\end{tabular}
		\end{center}
	\end{table}
\end{quote}

\ex Let $\eta = (+, +, -, -, -, +)$. This vector has a repeating block structure, $\eta = (\sigma,-\sigma)$, where $\sigma = (+,+,-)$. For $p=6$ the matrix $V$ is given by
$$
	V = (v^{(0)},v^{(1)},\dots,v^{(5)}) = \left(\begin{array}{rrrrrr} 
		1 &   1    &   1    &  1 &   1    &   1    \\
		1 & \rho   & \rho^2 & -1 & \rho^4 & \rho^5 \\
		1 & \rho^2 & \rho^4 &  1 & \rho^2 & \rho^4 \\
		1 &  -1    &   1    & -1 &   1    &  -1    \\
		1 & \rho^4 & \rho^2 &  1 & \rho^4 & \rho^2 \\
		1 & \rho^5 & \rho^4 & -1 & \rho^2 & \rho
	\end{array}\right),
$$
where $\displaystyle{\rho = e^{2\pi\mathbf{i}/6}}$. The polynomial $p_{\eta}(x)$ has the following factorization,
$$
	\begin{array}{lcl}
		p_{\eta}(x) & = & 1 + x - x^2 - x^3 - x^4 + x^5 		\\
		            & = & (1 - x^3)(1 + x - x^2)						\\
		            & = & (1 - x)(1 + x + x^2)(1 + x - x^2).
	\end{array}
$$
Since $\Phi_1(x) = x-1$ and $\Phi_3(x) = x^2 + x +1$ are the first and the third cyclotomic polynomials, and $\rho^0 = 1$ is the primitive first root of unity and $\rho^2$ and $\rho^4$ are the primitive third roots of unity, it follows that $\eta$ annihilates $v^{(0)}$, $v^{(2)}$ and $v^{(4)}$. Therefore, $N_6(\eta) = 6 - 3 = 3$.

Some of the $N$-values in $N_p(\mathbb{X}^p)$ in Table 1 can be explained directly, without factorizing $p_{\eta}(x)$. We summarize three simple but important facts.

\prop 

\noindent (a) $\{1,p\}\subset N_p(\mathbb{X}^p)$ for any $p>2$.

\noindent (b) If $p>2$ is prime, then $N_p(\mathbb{X}^p) = \{1,p\}$.

\noindent (c) $p-1\in N_p(\mathbb{X}^p)$ if $p$ is even and $p>4$.

\begin{proof}
		Since $\mathrm{rank}(+, +, \dots, +) = 1$, it follows that $1\in N_p(\mathbb{X}^p)$ for any $p$. To show that $p\in N_p(\mathbb{X}^p)$ for $p>2$, consider $\eta = (-, +, +, \dots, +)$ and let $\Sigma_{\eta}$ be the $p\times p$-matrix defined by $\mathrm{row}_j(\Sigma_{\eta}) = \eta(\mathbf{P}^T)^{j-1}$, $1\leq j\leq p$. By induction, one shows that $\det(\Sigma_{\eta}) = (-2)^{p-1}(p-2)$ which completes the proof of (a).

		Statement (b) is an immediate consequence of the fact that $\Phi_p(x) = \sum\limits_{j=1}^p x^{j-1}$ is the minimal polynomial of $\rho^k$, $0<k<p$, if $p$ is prime and is irreducible  over $\mathbb{Q}$, hence if $\eta \not= \pm(+,+,\dots,+)$, $\Phi_p(x)$ and $p_{\eta}(x)$ cannot contain a common factor.

		To show (c), let $\sigma = (+, -, +, -, \dots, +, -)\in\mathbb{X}^{p-2}$ and set $\eta = (\sigma, -, +)$. By performing elementary row operations on the matrix $\Sigma_{\eta}$ with rows $\mathrm{row}_i(\Sigma_{\eta})$ $ = \eta\mathbf{P}^{i-1}$, $1\leq i\leq p-1$, it can be shown that $\Sigma_{\eta}$ has full rank. The details are straightforward but tedious to write down explicitly and will be omitted.

\end{proof}

\rem 
\noindent (a) If $p$ is prime, then $\mathrm{rank}(\eta) = p$ for any $\eta \not= \pm(+,+,\dots,+)$. For non-prime values of $p$ one also can construct several different vectors with $\mathrm{rank}(\eta) = p$. For example, if $p$ is odd, then $\mathrm{rank}(\eta) = p$ if $\eta = (+,-,+,-,\dots,+,-,+)$, which is easily shown using elementary row operations. A generalization is provided by vectors $\eta$ with $\sum\limits_{i}\eta^i = 1$. All our case studies indicate that these vectors have $\mathrm{rank}(\eta) = p$ as well.

\noindent (b) For even $p>4$, the vector constructed in the proof of Proposition 8(c) is just one example of a vector with $\mathrm{rank}(\eta) = p-1$. In general, if $\eta = (\eta^1,\dots,\eta^p)$ and $\sum\limits_{i}\eta^i = 0$, then $\eta$ is orthogonal to $(+,+,\dots,+)$, and $\mathrm{rank}(\eta) \leq p-1$. Case studies indicate that such a vector has maximal rank $p-1$ if it does not have a ``repeating block structure''.

\begin{figure}
	\label{fig:F41NetTop}
	\begin{center}
		\includegraphics[height=3in]{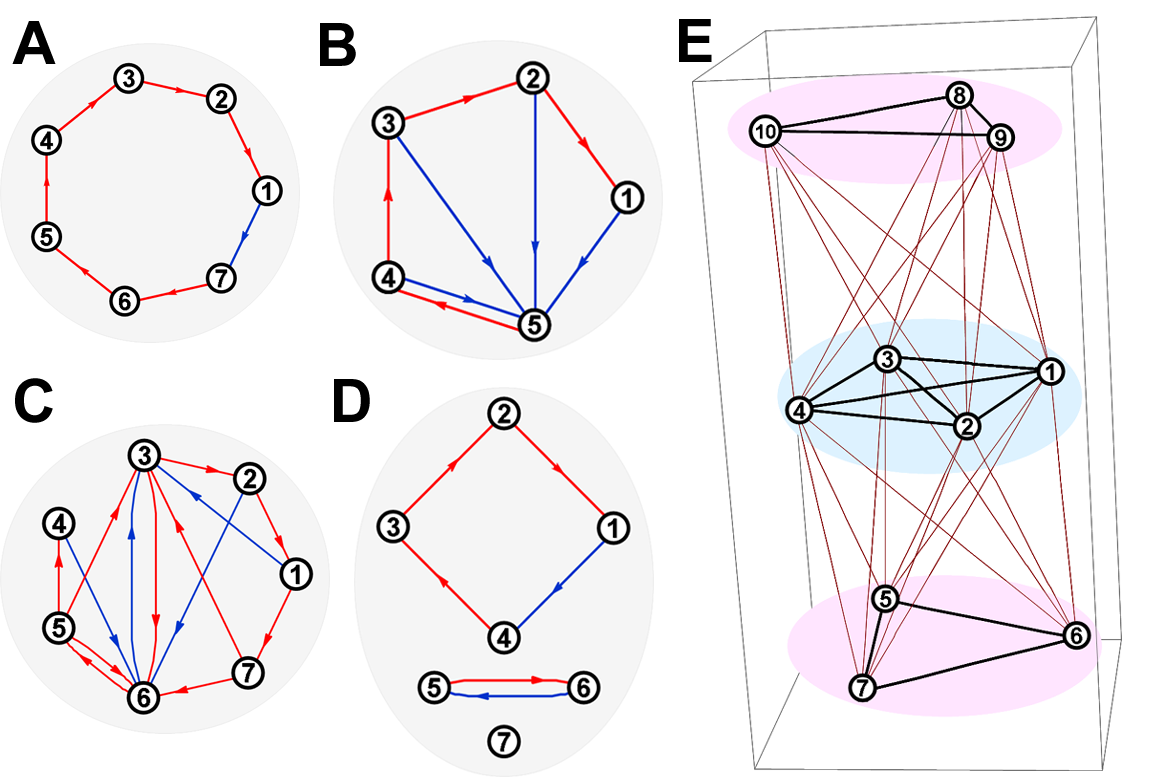}
	\end{center}
	\caption{Topology of networks constructed, respectively, by a simple MC-cycle with $N = p/2 = 7$ (\textbf{A}, Example 6), a simple MC-cycle with $N = p-1 = 5$ (\textbf{B}, Example 6), a minimal simple but non-consecutive cycle with $N = p-2 = 7$ (\textbf{C}, Example 7), a separable MC-cycle (\textbf{D}, Example 10) and a minimal genuinely inseparable composite cycle (\textbf{E}, Example 11). In panels \textbf{A-D}, excitatory (inhibitory) synaptic connections are labeled with red (blue) lines with arrowhead indicating the direction of the connections. In panel \textbf{E}, in order to highlight the clusters in the network, connections within the same clusters are labeled with black lines, and connections between neurons in different clusters are labeled with dark red lines. Directions and polarities of the connections are not shown in \textbf{E}. For all of the 5 networks illustrated in this figure, self-connectivities are not included.}
\end{figure}

\subsubsection{Simple anti-symmetric cycles}
The characteristics of the cycles considered by \cite{Gencic90} are that the cycle length $p$ is even and the second $p/2$ columns of the cycle matrix are the negatives of the first $p/2$ columns in the same order. We call such cycles \emph{anti-symmetric}. Here we discuss the possible values of the rank of the cycle matrix if these cycles are simple and admissible.

\prop Assume $p$ is even, $p=2n$, and $\eta = (\sigma, -\sigma)$ with $\sigma = (\sigma^1,\dots,\sigma^n)\in\mathbb{X}^n$. Then we have the following possibilities for $\mathrm{rank}(\eta)$.

\noindent (a) $1\leq\mathrm{rank}(\eta)\leq n$. Moreover, $d\equiv n-\mathrm{rank}(\eta)$ is even, and if $d\geq 2$ the $\sigma^j$ satisfy $d$ linearly independent homogeneous linear equations with integer coefficients.

\noindent (b) If $p=2^k$, $k\geq 2$, then $\mathrm{rank}(\eta) = n$.

\noindent (c) If $p=2n$ with $n > 2$ prime and $\sigma \not= \pm(+,-,+,\dots,+,-,+)$, then $\mathrm{rank}(\eta) = n$.

\begin{proof}
	\noindent (a) Let $\eta = (\sigma,-\sigma)$, $\sigma\in\mathbb{X}^n$, and define the $p\times p$-matrix $\Sigma$ by $\mathrm{row}_j(\Sigma) = \eta(\mathbf{P}^T)^{j-1}$, $1\leq j\leq p$. This matrix is a circulant matrix and contains all rows of $\mathcal{L}_{\eta}$, hence $\mathrm{rank}(\eta) = \mathrm{rank}(\Sigma)$. According to the properties of circulant matrices, the eigenvalues of $\Sigma$ are of the form
			\begin{equation}
				\label{eq:SigEig00}
				\lambda = \sum\limits_{i=1}^p \eta^i\rho^{i-1} = \sum\limits_{i=1}^n (\sigma^i - \sigma^i\rho^n)\rho^{i-1},
			\end{equation}
			where $\rho$ is any $p$th root of unity. Thus $\mathrm{rank}(\eta)$ coincides with the number of distinct $p$th roots of unity for which the right-hand side of (\ref{eq:SigEig00}) is nonzero. Since $\eta\mathbf{P}^n = -\eta$, it follows that $\mathrm{rank}(\eta) \leq n$, and clearly $\mathrm{rank}(\eta) \geq 1$, which proves the first statement of (a). 
			To complete the proof of (a), we note that the $2n$ distinct roots of $x^{2n} = 1$ ($x\in\mathbb{C}$) comprise $n$ roots with $x^n = 1$ and $n$ roots with $x^n = -1$. Thus $d = n - \mathrm{rank}(\eta)$ coincides with the number of distinct roots of $x^n = -1$ for which 
			\begin{equation}
				\label{eq:SigEig01}
				p_{\sigma}(x) = \sum\limits_{i=1}^n \sigma^i x^{i-1} = 0.
			\end{equation}
			If $n$ is odd, $p_{\sigma}(-1) \not= 0$, and if $n$ is even, $(-1)^n = 1$, thus all roots in question have nonzero imaginary parts, which implies that $d$ is even. If $d\geq 2$, $p_{\sigma}(x)$ is divisible by a cyclotomic polynomials $\Phi_m(x)$, where $m$ divides $n$ but not $2n$. The degree of $\Phi_m(x)$ is given by Euler's totient function, $\varphi(m)$, and is even. The condition that $p_{\sigma}(x)$ factors through $\Phi_m(x)$ then leads to $\varphi(m)$ linearly independent homogeneous equations that must be satisfied by the $\sigma^i$, and since $\Phi_m(x)$ has integer coefficients, the coefficients of these equations can be chosen as integers as well.
			If $p_{\sigma}(x)$ contains several cyclotomic polynomials $\Phi_{m_j}(x)$, $1\leq j\leq r$, as factors, the number of linear equations satisfied by $\sigma$ is $\varphi(m_1) + \cdots + \varphi(m_r)$, and all these relations are linearly independent as the cyclotomic polynomials are distinct and irreducible over the rationals.

	\noindent (b) If $p = 2^k$, $k\geq 2$, the only factor that divides $p$ but not $n = 2^{k-1}$ is $p$. The cyclotomic polynomial of $p$ is $\Phi_p(x) = 1 + x^n$ and has degree $n$, that is, $\Phi_p(x)$ cannot be a factor of $p_{\sigma}(x)$ which has degree $n-1$.

	\noindent (c) Assume now that $n>2$ is prime. In this case, $\Phi_{2n}(x) = \Phi_{n}(-x)$ is the only cyclotomic polynomial in question and is given by $\Phi_{n}(-x) = 1 - x + x^2 - x^3 + \cdots + x^n$. Thus, if $\sigma$ is not of the form $\sigma = \pm (+, -, +, -, \dots, +, -, +)$, $\Phi_n(-x)$ does not factor through $p_{\sigma}(x)$.
\end{proof}

\rm Since for $\eta = (\sigma,-\sigma)$ a ``rank deficiency'' ($\mathrm{rank}(\eta) < n$) occurs only if $\sigma$ satisfies a system of linear equations, the number of $\sigma$'s for which $\mathrm{rank}(\eta) = n$ is considerably larger than the number of $\sigma$'s for which $\eta$ has a rank-deficiency. Thus ``generically'' we expect that vectors of the form $(\sigma, -\sigma)$ have full rank $n$. The vector $\sigma = (+,-,+,-,\dots,+,-,+)$ is, of course, a very special case as $(\sigma, -\sigma)$ has the repeating block structure $(+,-,+,-,\dots,+,-)$ which has minimal rank 1. We illustrate the occurrence of rank deficiencies by an example.

\ex Let $p = 18 = 2\cdot 3^2$, i.e. $n=9$. The cyclotomic polynomials that can give rise to a rank deficiency are here $\Phi_6(x) = 1 - x + x^2$ and $\Phi_{18}(x) = 1 - x^3 + x^6$. The condition that $p_{\sigma}(x) = \sigma^1 + \sigma^2 x + \cdots + \sigma^9 x^8$ factors through $\Phi_6(x)$ leads to the equations
$$
	\begin{array}{lcc}
		\sigma^1 - \sigma^3 - \sigma^4 + \sigma^6 + \sigma^7 - \sigma^9 & = & 0, \\
		\sigma^2 + \sigma^3 - \sigma^5 - \sigma^6 + \sigma^8 + \sigma^9 & = & 0.
	\end{array}
$$
The only binary vector satisfying these conditions (up to cyclic permutations) are $\pm\sigma^{(1)}$, $\pm\sigma^{(2)}$ and $\pm\sigma^{(3)}$, where 
$$
	\begin{array}{lcl}
		\sigma^{(1)} & = & (+,+,+,+,+,+,+,-,+), \\
		\sigma^{(2)} & = & (+,+,+,-,+,+,-,-,+), \\
		\sigma^{(3)} & = & (+,+,-,+,+,-,+,-,+).
	\end{array}
$$
Since $\Phi_6(x)$ has a single pair of complex conjugate roots, $\mathrm{rank}(\sigma^{(\nu)},-\sigma^{(\nu)}) = 9 - 2 = 7$, $\nu = 1, 2, 3$. Similarly, in order that $\Phi_{18}(x)$ factors through $p_{\sigma}(x)$, the conditions $\sigma^j + \sigma^{j+3} = 0$ for $1\leq j\leq 6$ must be satisfied, leading to $\mathrm{rank}(\sigma,-\sigma) = 3$. All vectors with block structure $\sigma = (\tilde{\sigma},-\tilde{\sigma},\tilde{\sigma})$ with $\tilde{\sigma}\in \mathbb{X}^3$ have this property, and lead to $\eta = (\hat{\sigma}, \hat{\sigma}, \hat{\sigma})$ with $\hat{\sigma} = (\tilde{\sigma}, -\tilde{\sigma})$, i.e. $\mathrm{rank}(\eta) = \mathrm{rank}(\hat{\sigma})$. This includes $\tilde{\sigma} = (+,+,+)$ with $\mathrm{rank}(\hat{\sigma}) = 3$, and $\tilde{\sigma} = (+,-,+)$ with $\mathrm{rank}(\hat{\sigma}) = 1$. In the latter case, both $\Phi_6(x)$ and $\Phi_{18}(x)$ are factors of $p_{\sigma}(x)$.

The rank of generic vectors (without rank deficiency) of the form $(\sigma,\sigma)$ or $(\sigma,-\sigma)$ is equal to the length of $\sigma$. The converse question is under which circumstances a vector $\eta\in\mathbb{X}^p$ with $p$ even and $\mathrm{rank}(\eta) = p/2$ has this form. We state two simple sufficient conditions for this property.

\prop Assume $p=2n$, $\eta\in\mathbb{X}^p$, and $\mathrm{rank}(\eta) = n$. If $p = 2^k$, $k\geq 2$, or $n>2$ is prime, then $\eta$ is either of the form $(\sigma,\sigma)$ or $(\sigma, -\sigma)$ for some $\sigma\in\mathbb{X}^n$.
\begin{proof}
	Let $\eta = (\sigma,\hat{\sigma})$ with $\sigma$, $\hat{\sigma}\in\mathbb{X}^n$. We consider again the matrix $\Sigma$ defined in the proof of Proposition 9 with eigenvalues
	$$
		\lambda = \sum\limits_{i=1}^n (\sigma^i + \rho^n\hat{\sigma}^i)\rho^{i-1},
	$$
	where $\rho$ is a $p$th root of unity, $p=2n$. Assuming that $\mathrm{rank}(\eta) = n$, there exist precisely $n$ distinct roots $\rho$ of $x^{2n} = 1$ for which $\lambda = 0$. We decompose these roots again into roots satisfying $x^n = 1$ and $x^n = -1$, respectively, and set accordingly
	$$
		\lambda_{\pm}(x) = \sum\limits_{i=1}^n(\sigma^i \pm \hat{\sigma}^i)x^{i-1}.
	$$
	
\noindent (a) Assume that $p = 2^k$ ($n = 2^{k-1}$) for $k\geq 2$. If there exists a root $\rho$ of $x^n = -1$ for which $\lambda = 0$, $\lambda(x)$ must contain the cyclotomic polynomial $\Phi_{2n}(x) = 1 + x^n$ as a factor, which is only possible if $\sigma^i - \hat{\sigma}^i = 0$ for all $i$, because $\lambda(x)$ has at most degree $n-1$. Thus $\eta = (\sigma,\sigma)$ in this case. Conversely, assume that all roots $\rho$ for which $\lambda = 0$ are roots of $x^n = 1$. Then $\lambda_+(x)$ must contain all cyclotomic polynomials $\Phi_{\nu}(x)$ for $\nu = 1, 2, \dots, 2^{k-1}$ as factors. Since the product of these polynomials is $1 - x^n$, this cannot hold unless $\sigma^i + \hat{\sigma}^i = 0$ for all $i$, thus $\eta = (\sigma, -\sigma)$ in this case.
	
\noindent (b) The case $p=2n$ with $n>1$ prime is treated similarly. Here the cyclotomic polynomials which factor through $x^n - 1$ are $1-x$ and $\Phi_n(x) = 1 + x + \cdots + x^{n-1}$, and the cyclotomic polynomials which factor through $1+x^n$ are $1+x$ and $\Phi_n(-x)$. Since $n>1$, either $\Phi_n(x)$ is a factor of $\lambda_+(x)$ or $\Phi_n(-x)$ is a factor of $\lambda_-(x)$, which implies that either $\hat{\sigma} = \sigma$ or $\hat{\sigma} = -\sigma$.
\end{proof}

\rm An extension of Proposition 10 to more general values of $p$ appears highly nontrivial, because a multitude of cyclotomic polynomials have to be considered if the prime factorization of $n$ is more complicated. We have examined all vectors $\eta$ with $\mathrm{rank}(\eta) = p/2$ for $p\leq 20$ and found that all these vectors have the form $(\sigma,\sigma)$ or $(\sigma,-\sigma)$. Other vectors with $\mathrm{rank}(\eta) = p/2$ may exist for larger values of $p$, but if so we expect the number of these vectors to be much smaller than the number of $(\sigma,\sigma)$- or $(\sigma,-\sigma)$-vectors of full rank.

\subsection{Separable MC-Cycles}
For separable MC-cycles with generators $G_{\Sigma} = \{\eta_1,\eta_2,\dots,\eta_q\}$, the spaces $\mathrm{span}\{\mathcal{L}_{\eta_j}\}$ and $\mathrm{span}\{\mathcal{L}_{\eta_k}\}$ intersect trivially if $j\not=k$. If $\Sigma$ is in standard form, this implies immediately that $\mathbf{J}$ has a block structure, $\mathbf{J} = \mathrm{diag}(\mathbf{J}_1,\dots,\mathbf{J}_r)$, where $\mathbf{J}_k$ is an $N_k\times N_k$-matrix of the form (\ref{eq:J1Mini00}) with $N_k = \mathrm{rank}(\eta_k)$. Accordingly, a network constructed from a separable MC-cycle is decomposed into $r$ disconnected clusters and for each cluster the connectivity matrix has the form corresponding to a simple MC-cycle.

\ex Consider the $7\times 8$-cycle
$$
	\Sigma = \left(\begin{array}{rrrrrrrr}
		 + &  + &  + &  + &  - &  - &  - &  - \\
		 + &  + &  + &  - &  - &  - &  - &  + \\
		 + &  + &  - &  - &  - &  - &  + &  + \\
		 + &  - &  - &  - &  - &  + &  + &  + \\
		 + &  + &  - &  - &  + &  + &  - &  - \\
		 + &  - &  - &  + &  + &  - &  - &  +	\\
		 + &  - &  + &  - &  + &  - &  + &  -
	\end{array}\right).
$$
This cycle has the generators $\eta_1$, $\eta_5$, $\eta_7$ ($\eta_j = \mathrm{row}_j(\Sigma)$) and is separable and in standard form. Moreover, $\eta_4\mathbf{P} = -\eta_1$, $\eta_6\mathbf{P} = -\eta_5$, and $\eta_7\mathbf{P} = -\eta_7$. Thus the network is decomposed into three clusters consisting of neurons 1, 2, 3, 4, neurons 5, 6, and neuron 7, with cycle-connectivity matrices
$$
	\left(\begin{array}{rrrrrrrr}
		 0 &  1 &  0 &  0 \\
		 0 &  0 &  1 &  0 \\
		 0 &  0 &  0 &  1 \\
		-1 &  0 &  0 &  0
	\end{array}\right)\mbox{, }
	\left(\begin{array}{rrrrrrrr}
		 0 &  1 \\
		-1 &  0
	\end{array}\right),
$$
and $-1$, respectively. The topology of this network is illustrated in Figure 3D. We note, however, that the cluster consisting of neuron 7 cannot show oscillations without delay, since a $1D$ dynamical system does not have limit cycles. By contrast, with delay included, we can find oscillations already for $N=1$ for appropriate parameter values.

General separable cycles still can be decomposed into isolated clusters as $\mathrm{span}\{\mathrm{R}(\Sigma)\}$ is semisimple, however, the network topology in each cluster maybe more complicated (see Subsection 5.4). The issue with separable cycles is that, even if the subcycles corresponding to the different clusters are retrieved, these oscillations are in general not synchronized. We comment on this issue further in Section 6.

\subsection{Minimal Inseparable Cycles}
For minimal inseparable cycles $\Sigma$ with generators $G_{\Sigma} = \{\eta_1,\eta_2,\dots,\eta_q\}$, at least two subspaces $\mathrm{span}\{\mathcal{L}_{\eta_j}\cap\mathrm{R}(\Sigma)\}$ and $\mathrm{span}\{\mathcal{L}_{\eta_k}\cap\mathrm{R}(\Sigma)\}$ ($j \not= k$) have a nontrivial intersection. Accordingly, $\Sigma$ does not have full row-rank,
$$
	\mathrm{rank}(\Sigma) < \sum\limits_{i=1}^{q}\mathrm{rank}(\eta_i) = N,
$$
which implies in particular that $\mathbf{J^0}\not=I$. It is still possible to partition the network into clusters, but some clusters may be connected and the network topology within the cluster corresponding to the loop $\mathcal{L}_{\eta_j}$ will in general not coincide with the network topology predicted by the submatrix $\Sigma_j$ of the corresponding simple cycle. Thus the consecutiveness requirement does not have an effect, whereas the minimality requirement takes care that the sub-matrices in $\mathbf{J^0}$ and $\mathbf{J}$ defining the connectivities within the clusters are non-singular. The following example illustrates these features.

\ex Consider the $10\times 12$-cycle $\Sigma = (\Sigma_1^T,\Sigma_2^T,\Sigma_3^T)^T$, where 
$$
	\begin{array}{l}
		\Sigma_1^T = (\eta_1^T, (\eta_1\mathbf{P})^T, (\eta_1\mathbf{P}^2)^T, (\eta_1\mathbf{P}^3)^T)^T, \\
		\Sigma_j^T = (\eta_j^T, (\eta_j\mathbf{P})^T, (\eta_j\mathbf{P}^2)^T)^T\mbox{, }j=2,3,
	\end{array}
$$
with 
$$
	\begin{array}{l}
		\eta_1 = (\begin{array}{cccccccccccc} + & + & + & - & + & + & + & - & + & + & + & - \end{array}), \\
		\eta_2 = (\begin{array}{cccccccccccc} - & + & + & + & - & - & - & + & + & + & - & - \end{array}), \\
		\eta_3 = (\begin{array}{cccccccccccc} + & - & - & + & - & - & + & - & - & + & - & - \end{array}).
	\end{array}
$$
This is a minimal inseparable admissible cycle with generator $G_{\Sigma} = \{\eta_1,\eta_2,\eta_3\}$. The connectivity matrix $\mathbf{J}$ is given by
$$
	\mathbf{J} = \frac{1}{8}\left(\begin{array}{rrrrrrrrrr}
		 0 &  7 &  0 & -1 &  1 & -1 &  1 & -1 & -1 & -1 \\
		-1 &  0 &  7 &  0 & -1 &  1 & -1 & -1 & -1 & -1 \\
		 0 & -1 &  0 &  7 &  1 & -1 &  1 & -1 & -1 & -1 \\
		 7 &  0 & -1 &  0 & -1 &  1 & -1 & -1 & -1 & -1 \\
		 1 & -1 &  1 & -1 &  2 &  6 &  2 &  0 &  0 &  0 \\
		-1 &  1 & -1 &  1 & -2 &  2 &  6 &  0 &  0 &  0 \\
		 1 & -1 &  1 & -1 &  6 & -2 &  2 &  0 &  0 &  0 \\
		-1 & -1 & -1 & -1 &  0 &  0 &  0 & -2 &  6 & -2 \\
		-1 & -1 & -1 & -1 &  0 &  0 &  0 & -2 & -2 &  6 \\
		-1 & -1 & -1 & -1 &  0 &  0 &  0 &  6 & -2 & -2
	\end{array}\right),
$$
and $\mathbf{J^0}$ has the same block structure as $\mathbf{J}$ (with self-feedbacks of all neurons).
From the form of $\mathbf{J}$ (and $\mathbf{J^0}$) we infer that the cluster corresponding to $\eta_1$ is connected to the clusters corresponding to $\eta_2$ and $\eta_3$, while the latter two clusters are not directly connected. This connectivity structure is due to the fact that $\mathrm{span}\{\mathcal{L}_{\eta_1}\}$ intersects $\mathrm{span}\{\mathcal{L}_{\eta_2}\}$ and $\mathrm{span}\{\mathcal{L}_{\eta_3}\}$ in the one-dimensional spaces spanned by $(+,-,+,-,\dots,+,-)$ and $(+,+,+,\dots,+)$, respectively, whereas $\mathrm{span}\{\mathcal{L}_{\eta_2}\}$ and $\mathrm{span}\{\mathcal{L}_{\eta_3}\}$ intersect trivially. The network topology for this example is shown in Figure 3E. In general, two clusters corresponding to two generators $\eta,\eta'\in G_{\Sigma}$ are connected, if there exists a sequence $\eta = \eta_1, \eta_2, \dots, \eta_{s-1}, \eta_s = \eta'$ of generators such that $\mathrm{span}\{\mathcal{L}_{\eta_j}\}$ and $\mathrm{span}\{\mathcal{L}_{\eta_{j+1}}\}$ intersect nontrivially for $0\leq j<s$.

\subsection{Further Examples}
The examples in this subsection serve to illustrate the possible effects of fewer essential generators than generators. Consider a cycle $\Sigma$ with $EG_{\Sigma} = \{\epsilon_1,\dots,\epsilon_r\}$. If $r<|G_{\Sigma}|$, the loop vectors of at least one generator are contained in the span of the loop vectors of another essential generator. Assuming $|\mathrm{R}(\Sigma)| = N$, this implies
$$
	\mathrm{rank}(\Sigma) \leq \sum\limits_{i=1}^r\mathrm{rank}(\epsilon_i) < N,
$$
and we encounter again a rank-deficiency that will destroy special structures in the clusters corresponding to the essential generators.

\ex The $6\times 6$-cycle
$$
	\Sigma = \left(\begin{array}{cccccc}
		+ & + & - & - & + & - \\
		+ & - & - & + & - & + \\
		- & - & + & - & + & + \\
		- & + & - & + & + & - \\
		+ & - & + & + & - & - \\
		+ & - & + & - & + & -
	\end{array}\right),
$$
has two generators, $\mathrm{row}_1(\Sigma)$ and $\mathrm{row}_6(\Sigma)$, but $\mathrm{row}_6(\Sigma) = \mathrm{row}_1(\Sigma) + \mathrm{row}_3(\Sigma) + \mathrm{row}_5(\Sigma)$, thus there is only one essential generator, $\epsilon = \mathrm{row}_1(\Sigma)$. Without the sixth row, $\Sigma$ would be a simple MC-cycle with ring-topology. The presence of the sixth row destroys this structure, which is revealed in the following forms of $\mathbf{J^0}$ and $\mathbf{J}$,
$$
	\mathbf{J^0} = \frac{1}{4}\left(\begin{array}{rrrrrr}
		 3 &  0 & -1 &  0 & -1 &  1 \\
		 0 &  4 &  0 &  0 &  0 &  0 \\
		-1 &  0 &  3 &  0 & -1 &  1 \\
		 0 &  0 &  0 &  4 &  0 &  0 \\
		-1 &  0 & -1 &  0 &  3 &  1 \\
		 1 &  0 &  1 &  0 &  1 &  3
	\end{array}\right)\mbox{, }\mathbf{J} = \frac{1}{4}\left(\begin{array}{rrrrrr}
		 0 &  4 &  0 &  0 &  0 &  0 \\
		-1 &  0 &  3 &  0 & -1 &  1 \\
		 0 &  0 &  0 &  4 &  0 &  0 \\
		-1 &  0 & -1 &  0 &  3 &  1 \\
		-1 & -4 & -1 & -4 & -1 & -3 \\
		-1 &  0 & -1 &  0 & -1 & -3
	\end{array}\right).
$$

\ex The cycle
$$
	\Sigma = \left(\begin{array}{rrr}
		 1 & -1 &  1 \\
		-1 &  1 & -1 \\
		 1 &  1 &  1
	\end{array}\right),
$$
has three generators and one essential generator that can be chosen as the first or second row. Since $\Sigma$ is non-singular, $\Sigma$ is admissible and $\mathbf{J^0}$ is the identity matrix. A successfully retrieved cycle shows three consecutive phases 1, 2, and 3 during an oscillation. In phases 1 and 2, neuron 1 is ``on'' ($+$) and in phase 3 it is ``off'' ($-$), while neuron 2 is ``on'' in phase 1 and ``off'' in phases 2 and 3. Clearly, neuron 3 is ``on'' during all 3 phases. The matrix $\mathbf{J}$ is given by
$$
	\mathbf{J} = \left(\begin{array}{rrr}
		-1 &  1 &  1 \\
		-1 &  0 &  0 \\
		 0 &  0 &  1
	\end{array}\right),
$$
and shows that neurons 1 and 2 form an excitatory/inhibitory pair, whereas neuron 3 acts excitatory on neuron 1. Without this third neuron the oscillations of neurons 1 and 2 as required by the first two rows of $\Sigma$ could not be implemented, since the submatrix of $\Sigma$ consisting of these rows is not admissible.

\section{Discussion and Conclusion}
In this paper we have studied the structural features of admissible cycles and their relation to the topology of the corresponding networks. While our main motivation was the storage of cycles in continuous-time Hopfield-type networks, the results apply to other networks as well, including the discrete networks considered by \cite{Personnaz} and \cite{Guyon} and networks of spiking neurons exhibiting up-down states. In particular, we have formulated and proved conditions on binary cyclic patterns that guarantee the existence of a network with connectivity satisfying the transition conditions imposed by the cycle, independent of the specific dynamics of the individual neurons.

We showed that if and only if the discrete Fourier transform $\hat{\Sigma} = \Sigma V$ of a cycle matrix $\Sigma$ contains exactly $r$ nonzero columns, where $r = \mathrm{rank}(\Sigma)$, then a network can be constructed from $\Sigma$ with the pseudoinverse learning rule. Based on the structural analysis of the invariant subspaces of the row space of $\Sigma$, the admissible cycles have been classified into simple cycles, and separable and inseparable composite cycles. This classification was based on the decomposition of the row space of $\Sigma$ into subsets corresponding to disjoint loops. The admissibility of a cycle implied that all vectors of a loop are in the row space of $\Sigma$ if $\Sigma$ contains some of these loop vectors. If no loop-space associated with $\Sigma$ is a subspace of another loop-space (the generators are essential generators), we have identified for each loop the neurons associated with the loop vectors contained in $\Sigma$ with a cluster. For general admissible cycles the clusters are connected, and the connectivity of the clusters depends on the intersections of their loop-spaces. Two clusters are directly connected if their indecomposable invariant subspaces intersect non-trivially. They are ``indirectly'' connected if they are part of a chain of directly connected clusters.

If an admissible cycle is separable, the clusters are completely isolated. In this case each cluster corresponds to a simple cycle associated with a generator of $\Sigma$. If the simple cycle is minimal and consecutive, the cluster has the form of a feedforward chain from the last neuron to the first neuron with feedbacks to the last neuron from the other neurons. If in addition the length of the cycle, $p$, is even and the rank of the generator is $p/2$, we generically find a ring structure with excitatory or inhibitory connection from the first neuron to the last neuron, but we cannot exclude that special loops with these properties exist for which no ring-structure occurs. If the simple cycle is minimal but non-consecutive, we find more than one feedforward chains. 

Regarding non-minimal simple as well as composite cycles, it would be interesting to find equivalence relations similar to those of \cite{Golubitsky05}, and \cite{Golubitsky06}, relating networks constructed from non-minimal cycles to networks constructed from minimal cycles. For example, similar to the linear-threshold (LT) networks \citep{Tang06,Tang10}, we may consider the Hopfield-type network (\ref{eq:HopfieldNetwork05}) with delay in a different but closely related form,
\begin{equation}
	\label{eq:HopfieldNetworkDefnScalarNew}
	\mathbf{\dot{u}} = -\mathbf{u} + \tanh\left(\beta\left(C_0\mathbf{J^0}\mathbf{u} + C_1\mathbf{J}\mathbf{u}_{\tau}\right)\right),
\end{equation}
where $\beta = \beta_K\lambda$ and $\mathbf{u}_{\tau} = \mathbf{u}(t - \tau)$. While (\ref{eq:HopfieldNetwork05}) is invariant under arbitrary permutations of the neurons, it can be shown that (\ref{eq:HopfieldNetworkDefnScalarNew}) is invariant under a larger class of linear transformations that allows to define broad equivalence relations among admissible cycles. In comparison to (\ref{eq:HopfieldNetwork05}), the only disadvantage (\ref{eq:HopfieldNetworkDefnScalarNew}) may have is that it is less biologically plausible, because in biological neural networks neurons usually are coupled with each other through chemical synapses, which means that the firing rates instead of the membrane potentials of the presynaptic neurons change the membrane potential of the postsynaptic neuron.

In networks constructed from composite cycles, the complete isolation of the clusters of separable cycles means that each cluster has its own subcycle. The issue is that we cannot expect the different subcycles to synchronize, preventing the network to traverse the cycle states in the order prescribed by the cycle matrix. In this case an additional synchronization mechanism must be introduced to enforce synchrony. Such a mechanism can be in the form of a small coupling among the clusters or through an external periodic input acting as pacemaker.

The generation of cyclic patterns in animal nervous systems is associated with CPG networks, and the storage and retrieval of cyclic patterns in such networks are fundamentally important. Recent experimental observations \citep[e.g.][]{Dickinson92,Meyrand94,Jean01} suggested that CPGs may be highly flexible. As some animal movements, such as swallowing, gastrointestinal motility etc., often require the coordination of several functional groups of muscles, different CPGs controlling these muscles subsequently form during different phases of the movements. Such CPG networks consist of pools of neurons that can function in several CPGs involved in the organization of various motor behavior. 

Recently, in order to account for the flexibility of memory representation observed in neurophysiological experiments, \cite{Tang10} studied the effect of saliency weights on the memory dynamics in LT neural networks. They showed that the saliency distribution determines the retrieval process of the stored patterns, and that a nonuniform saliency distribution can contribute to the disappearance of spurious states. Using our results on the relation between the structural features of a cycle and the network topology, a mechanism similar to the variable saliency factor introduced by \cite{Tang10} into LT networks may be used to combine different CPGs in one network, and to study how a sequence of several cycles determines a changing network structure.

\section*{Acknowledgement}
The authors would like to thank the anonymous reviewers for their valuable comments and suggestions, and Alexander Hulpke and Chuck Anderson for helpful discussions. The first author was supported in part by a Summer Graduate Research Fellowship (2012) awarded by the Department of Mathematics at Colorado State University.

\appendix\section{Cyclotomic Polynomials}
We summarize here the basic properties of the cyclotomic polynomials used in Sections 3-5, for details see \cite{DummitFoote}.

The cyclotomic polynomial of order $p$ is defined by $\Phi_p(x) = \prod_r(x-x_r)$, $x\in\mathbb{C}$, where the $x_r$ encompass all primitive $p$-th roots of unity, that is, $x_r^p = 1$ and $x_r^n\not=1$ if $1\leq n<p$. The total number of such primitive roots is given by \emph{Euler's totient function}, $\varphi(p)$. If $p = \prod_j p_j^{m_j}$ with distinct primes $p_j$ is the prime factorization of $p$, then $\varphi(p) = \prod_j p_j^{m_j-1}(p_j-1)$. The important property of the cyclotomic polynomial is that they have integer coefficients and are irreducible over the rationals. Moreover, $\Phi_p(x)$ is the minimal polynomial for each root $x_r$, and the product of all $\Phi_d(x)$ for which $d$ is a factor of $p$ and $1\leq d\leq p$ is $x^p - 1$. The only cyclotomic polynomials of odd degree are $\Phi_1(x) = 1-x$ and $\Phi_2(x) = 1+x$, all $\Phi_p(x)$ for $p>2$ have even degrees as their primitive roots are all complex. Some basic properties of $\Phi_p(x)$ are:
	$$
		\Phi_p(x) = \sum\limits_{i=1}^{p}x^{i-1}\mbox{ if }p\mbox{ is prime,}
	$$
	$$
		\Phi_{2p}(x) = \Phi_p(-x)\mbox{ if }p\mbox{ is odd,}
	$$
	$$
		\Phi_p(x) = \Phi_q(x^{p/q}),
	$$
	where $q$ is the radical of $p$, i.e. the product of all distinct prime numbers occurring in the prime factorization of $p$. The last property implies in particular $\Phi_p(x) = 1 + x^n$ if $p = 2n = 2^k$, $k \geq 1$. The third cyclotomic polynomial is $\Phi_3(x) = 1 + x + x^2$ and has the roots $\displaystyle{e^{\pm2\pi\mathbf{i}/3}}$.





\section*{References}


\end{document}